\documentclass{article}
\usepackage[T1]{fontenc}
\usepackage[preprint]{neurips_2026}
\usepackage{times}

\usepackage{amsmath}
\usepackage{amssymb}
\usepackage{booktabs}
\usepackage{graphicx}
\usepackage{hyperref}
\usepackage{url}
\usepackage{xcolor}

\newif\ifrevisionmarkup
\ifdefined\showrevisionmarkup\revisionmarkuptrue\fi
\newcommand{\revadd}[1]{\ifrevisionmarkup{\color{blue!70!black}#1}\else#1\fi}
\newcommand{\revdel}[1]{\ifrevisionmarkup{\color{red!70!black}\noindent\textbf{[Original]} #1\par}\fi}
\newcommand{\revnote}[1]{}
\newcommand{\revnew}[1]{\ifrevisionmarkup{\color{blue!70!black}#1}\else#1\fi}
\newcommand{\revold}[1]{\ifrevisionmarkup{\color{red!70!black}\noindent\textbf{[Original]} #1\par}\fi}

\title{Beyond Successful Play: Diagnosing Opponent Modeling in Large Language Models}

\author{%
Hanbo Xie\thanks{Equal contribution. Corresponding author: \texttt{hxie88@gatech.edu}.} \\
Georgia Institute of Technology
\And
Yue Zhao\footnotemark[1] \\
University of Sussex
\And
Robert C. Wilson \\
Georgia Institute of Technology
}

\begin{document}

\maketitle
\ifrevisionmarkup
\begin{center}
\footnotesize\textcolor{blue!70!black}{Blue: current wording.}\quad
\textcolor{red!70!black}{Red: wording before the first marked front-matter revision.}
\end{center}
\fi

\begin{abstract}
Language-model agents are expected to adapt to others whose behavior changes through interaction. Yet adaptation may rely on learned behavioral patterns rather than theory of mind (ToM): the ability to infer others' mental states, including beliefs and intentions. We introduce a benchmark and diagnostic framework for repeated hide-and-seek against algorithmic opponents with explicitly defined recursive belief updates. Their ground-truth beliefs and action probabilities let us distinguish adaptive play, action prediction, and belief tracking through live interaction, fixed-history evaluation, and controlled information interventions. With structured elicitation and behavioral supervised fine-tuning (SFT), we find improved play in several tested settings, but diagnostics reveal limitations in opponent prediction and belief tracking. We test whether supervising opponent action probabilities can address this predictive deficit in Qwen2.5-7B-Instruct. Predictions improve when the model is told the opponent's type and parameter settings, but become less accurate when it must infer the opponent from history without this information. Training on examples that also omit these details reduces the gap, although improvements vary across opponents. Is this limitation specific to smaller open-weight models, or does it also appear in a frontier model? GPT-5.4 also benefits from correct opponent beliefs: supplying them improves its predictions and choices, whereas deeper explicit belief reports and self-generated memory do not consistently improve forecasting from interaction history. This contrast suggests that using useful belief information is easier than inferring it autonomously, even for a frontier model. Together, our findings distinguish improved play and action prediction from reliable belief tracking, providing a framework for evaluating mentalizing and testing targeted improvements.
\end{abstract}

\section{Introduction}
\label{sec:introduction}

\revdel{Large language models (LLMs) are increasingly evaluated on their ability to reason about others' minds, often through fixed narratives and questions about characters' beliefs \citep{kosinski2024evaluating,vanduijn2023theory}. These evaluations target theory of mind (ToM), or mentalizing: the capacity to infer others' latent mental states, including their beliefs, intentions, and goals \citep{premack1978chimpanzee}. However, performance on a particular task is not a complete measure of this capacity. Cognitive research emphasizes that false-belief tasks impose demands beyond mental-state reasoning and capture only one aspect of ToM \citep{bloom2000falsebelief}. It also distinguishes understanding mental-state concepts from deploying them effectively and spontaneously in context \citep{apperly2012theory}. These distinctions raise an important question for LLM evaluation: when a model answers correctly or acts successfully, what evidence supports interpreting that success as mentalizing?}
\revadd{Large language models (LLMs) are often evaluated for theory of mind (ToM), or mentalizing, through questions about characters' beliefs in fixed narratives \citep{kosinski2024evaluating,vanduijn2023theory}. Mentalizing concerns inference about another agent's latent beliefs, intentions, and goals \citep{premack1978chimpanzee}. An expected answer on a familiar task does not rule out learned response patterns or shortcut cues, and small changes to a scenario can alter performance \citep{ullman2023trivial}. When a model answers correctly or acts successfully, what evidence supports interpreting that success as mentalizing?}

\revnote{Figure 1 is placed at a page top under the official template.}
\begin{figure}[!t]
    \centering
    \includegraphics[width=0.325\textwidth,height=0.325\textwidth]{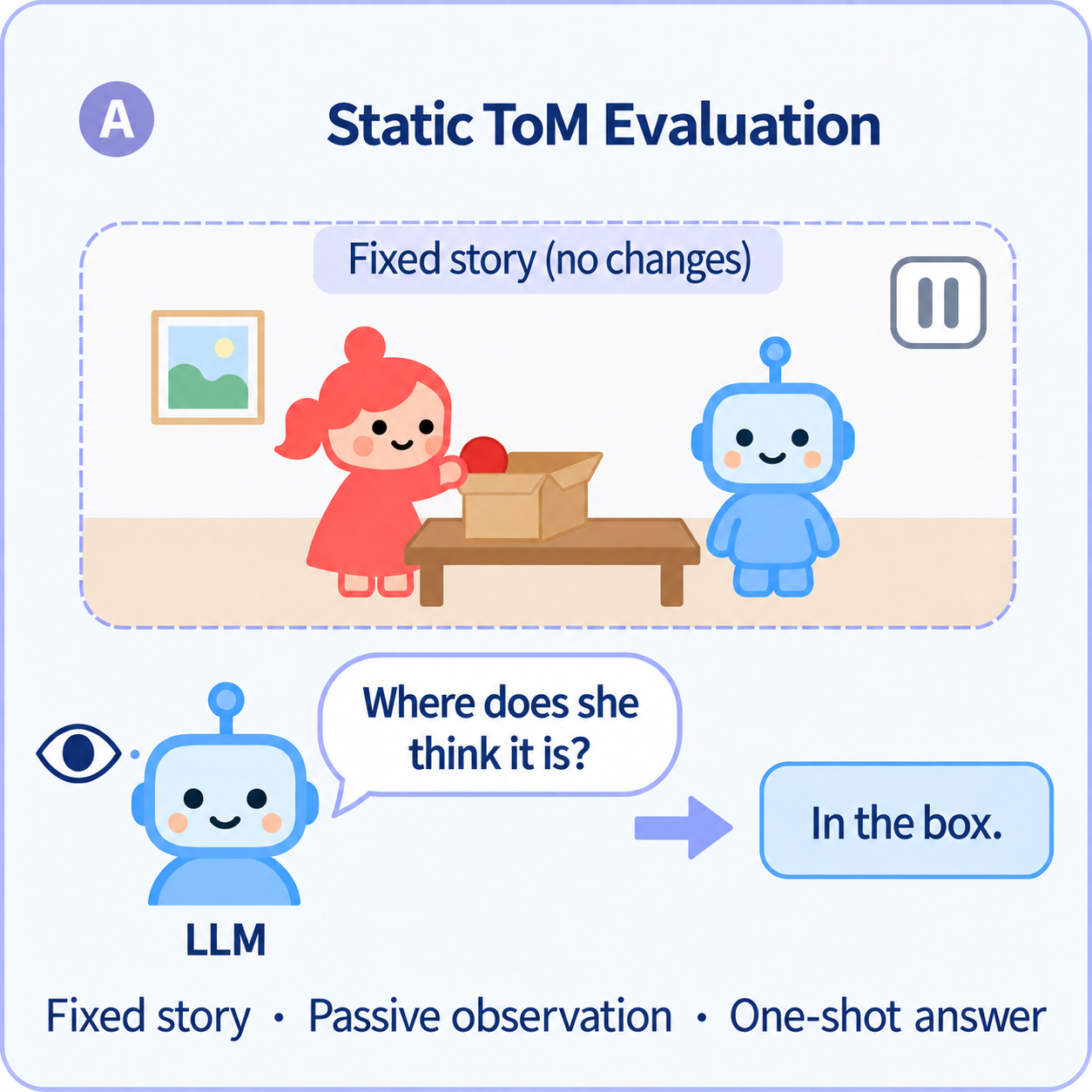}
    \hfill
    \includegraphics[width=0.325\textwidth,height=0.325\textwidth,trim=7bp 8bp 7bp 7bp,clip]{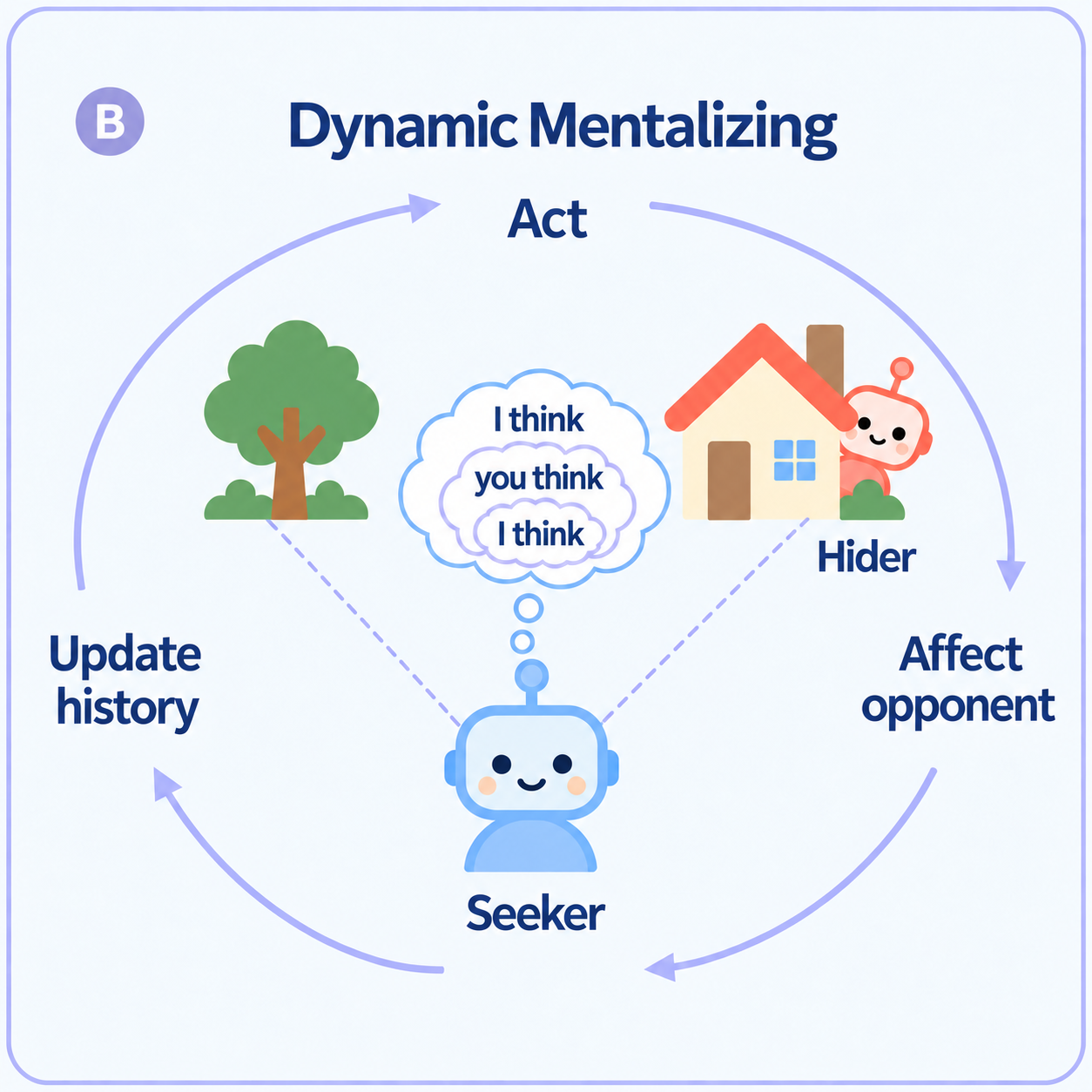}
    \hfill
    \includegraphics[width=0.325\textwidth,height=0.325\textwidth,trim=23bp 27bp 24bp 35bp,clip]{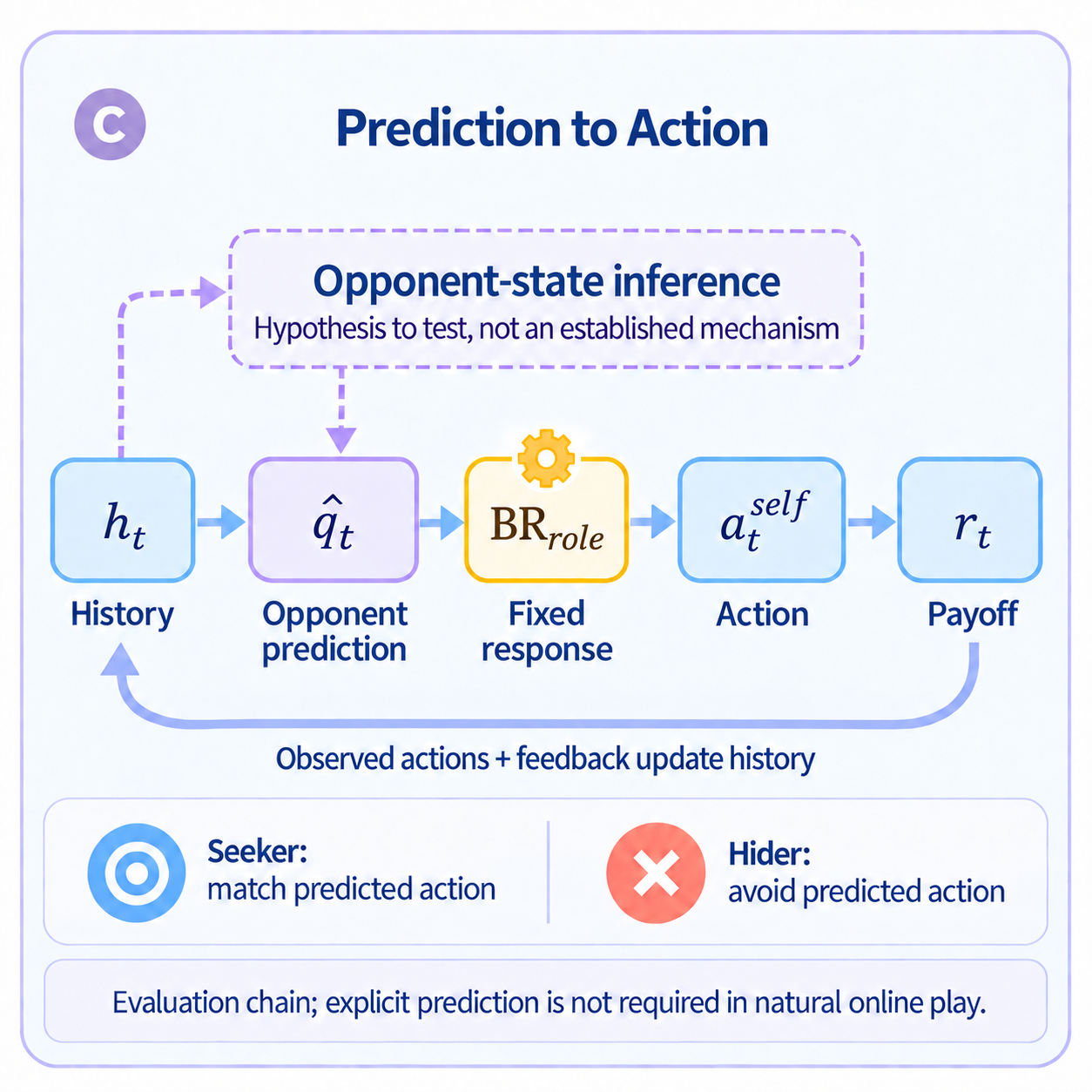}

    \caption{
    \textbf{From static ToM evaluation to dynamic opponent modeling.}
    (A) Static evaluations ask models to answer belief questions
    about a fixed narrative.
    (B) Repeated interaction requires anticipating an adaptive
    opponent while one's own actions shape subsequent observations.
    (C) Our evaluation distinguishes opponent action prediction,
    role-conditioned response, and payoff.
    The dashed pathway denotes opponent-state inference as a
    process to investigate, not an established internal mechanism.
    Explicit prediction is required only in designated evaluation
    interfaces, not in natural online play.
    }
    \label{fig:task-schematic}
\end{figure}

\revdel{This question becomes particularly important in strategic interaction. A successful choice may arise from exploiting behavioral regularities or from reasoning about the beliefs guiding another player's actions. The latter can become recursive when the other player is also forming predictions about oneself \citep{devaine2014evolution}. As both players respond to the interaction, their choices change not only immediate outcomes but also the evidence available for subsequent inference. Recent LLM evaluations have begun to address active planning and repeated strategic interaction, including cognitive-model analyses of observed choices \citep{slater2026planning,sohail2026mentalization}. These approaches provide complementary evidence, but effective play does not uniquely establish that a model has inferred the mental states guiding its opponent's behavior. Our goal is therefore to examine whether improvements in play and prediction are accompanied by an ability to infer and use information about the opponent's internal states.}
\revnew{Fixed belief questions hold the evidence constant. In repeated interaction, each action changes both the payoff and the observations available at the next decision, and an opponent that predicts the model can adapt in turn \citep{devaine2014evolution}. Thus, success on static questions does not establish online adaptation, while live payoff alone does not establish belief inference: a model may exploit action frequencies or reward regularities. The sharper test is whether a model can infer useful opponent information from interaction history, rather than merely use it when supplied. Recent LLM studies examine planning and repeated strategic games \citep{slater2026planning,sohail2026mentalization}. We build on them by assessing payoff alongside direct action forecasts and controlled access to opponent information.}

\revold{We investigate these questions in a repeated two-action hide-and-seek game. The seeker benefits from matching the opponent's action, whereas the hider benefits from choosing a different action. This minimal task limits demands from action search while retaining the challenge of anticipating an adaptive opponent: each player's actions become evidence that can shape the other's subsequent behavior. We evaluate models against algorithmic opponents ranging from simple behavioral strategies to recursive Bayesian $k$-ToM agents implemented with \texttt{tomsup} \citep{waade2023tomsup}. Because these opponents have explicitly defined update and decision rules, the simulator provides reference action probabilities and internal states that support evaluation beyond realized outcomes. These quantities are hidden from the evaluated model by default and supplied only in designated diagnostic conditions. Figure~\ref{fig:task-schematic} summarizes the transition from static belief questions to dynamic interaction and the evaluation framework.}
\revnew{We study this distinction in a repeated two-action hide-and-seek game. The seeker benefits from matching the opponent's action, whereas the hider benefits from choosing a different action. This minimal task limits action-search demands while retaining the challenge of anticipating an adaptive opponent. Simple behavioral strategies provide controls for predicting action regularities; in Bayesian $k$-ToM opponents, $k$ indexes the order of reasoning about other agents, and higher orders have explicit belief updates. These opponents are implemented with \texttt{tomsup} \citep{waade2023tomsup}. The simulator provides reference next-action probabilities and, where defined, reference beliefs. We use \emph{opponent information} for the broader set of opponent characteristics and simulator states. The \emph{opponent specification} is the narrower bundle of type, parameters, state schema, and public initialization; it excludes the current state and exact update rules. Natural play withholds both. Designated diagnostics supply selected information. Figure~\ref{fig:task-schematic} summarizes the task and evaluation.}

To organize the evaluation, we consider a functional prediction-to-action chain:
\begin{equation}
h_t
\longrightarrow \hat q_t
\longrightarrow \mathrm{BR}_{\mathrm{role}}(\hat q_t)
\longrightarrow a_t^{\mathrm{self}}
\longrightarrow r_t.
\label{eq:functional-chain}
\end{equation}
\revdel{Here, the interaction history $h_t$ provides evidence for predicting the opponent's next action.}
\revadd{For the round-$t$ decision, $h_t=\{(a_s^{\mathrm{self}},a_s^{\mathrm{op}},r_s)\}_{s=1}^{t-1}$ contains both players' realized actions and the focal player's payoff from completed rounds. The model's forecast $\hat q_t$ estimates $q_t$, the simulator's probability that the opponent chooses action 1 at $t$.} A role-conditioned best response translates that forecast into a preferred choice: matching the opponent's more likely action for the seeker, or avoiding it for the hider. The deployed action, together with the opponent's realized action, determines the round payoff. Observed actions and feedback then extend the history to $h_{t+1}$, closing the interaction loop. This decomposition separates forecast quality from choices and outcomes. Natural online play need not produce an explicit prediction or reasoning report.

\revold{Our benchmark combines live interaction, fixed-history prediction, and controlled information comparisons. Live interaction measures performance when the model's actions contribute to the evolving history. Fixed-history evaluation compares predictions on shared recorded trajectories, without allowing the evaluated model to alter future observations. Controlled information comparisons test whether models benefit from opponent characteristics or simulator beliefs that they would otherwise need to infer from behavior. Our primary evaluations concern payoff and opponent action prediction, while selected state-based diagnostics examine the information supporting these outcomes. We do not assume that effective mentalizing requires reproducing every simulator state or generating an explicit verbal reasoning chain.}
\revnew{We combine three evaluations. Live interaction measures payoff as the model's actions shape future histories. Fixed-history tests score forecasts on shared recorded trajectories. Information interventions supply opponent type and parameters or simulator-defined beliefs that are hidden during natural play. These comparisons separate payoff, action prediction, and use of supplied information. Selected belief-state diagnostics compare reports with simulator targets. None of these measures alone identifies the model's internal algorithm, nor do we require natural play to produce an explicit reasoning chain.}

\revold{Our experiments show that structured elicitation and fine-tuning can improve play and opponent prediction in the tested settings. However, these gains remain sensitive to the information available to the model. Predictive fine-tuning is more effective when opponent information is provided, while complementary diagnostics with a frontier model show benefits from supplied opponent beliefs that are not consistently obtained through explicit belief reporting or self-generated memory. Together, these findings highlight a distinction between using useful opponent information and inferring it from interaction history.}
\revnew{Structured elicitation and SFT improve some payoff and forecast measures, but gains depend on information access. Qwen2.5-7B fine-tuning on opponent-action probability performs better with opponent specification; when it is withheld, improvements vary across opponents. With GPT-5.4, supplying simulator beliefs improves predictions and choices, whereas explicit belief reports and self-generated memory do not consistently improve forecasts from history. Across these separate tests, using supplied information is easier to demonstrate than inferring it from interaction.}

Our contributions are twofold:
\begin{itemize}
    \item \revold{\textbf{A controlled benchmark for dynamic opponent modeling.} We evaluate language models in repeated strategic interaction against adaptive algorithmic opponents with accessible reference action probabilities and internal states. Combining live interaction, fixed-history prediction, and controlled information comparisons allows us to evaluate payoff and opponent prediction separately, and examine the information supporting them.}\revnew{\textbf{A controlled benchmark.} Against opponents with known update rules, we assess live payoff, action forecasts, and selected belief reports separately.}
    \item \revold{\textbf{Evidence distinguishing the use of opponent information from its inference.} Across predictive fine-tuning experiments and frontier-model diagnostics, models benefit from supplied opponent information, but these gains do not consistently carry over when that information must be inferred from interaction history. These findings reveal an important boundary of improvements in play and prediction: better performance alone does not establish reliable inference of an opponent's internal states.}\revnew{\textbf{A test of inference versus use.} Probability-supervised training and information interventions ask whether gains persist when opponent information must be inferred from history.}
\end{itemize}

\section{Related Work}
\label{sec:related-work}

\subsection{Evaluating Mentalizing in Large Language Models}

\revdel{False-belief tasks are a common starting point for evaluating ToM in LLMs. These tasks typically present a narrative in which a character has incomplete or outdated information, then ask what that character believes or how they will act. Studies using false-belief and related social-reasoning tasks have reported substantial capabilities in some models, alongside sensitivity to task formulation and alterations of the scenario \citep{kosinski2024evaluating,vanduijn2023theory,ullman2023trivial}.}

\revdel{Interpreting these results requires distinguishing task performance from the broader capacity it is intended to measure. In cognitive research, \citet{bloom2000falsebelief} argue that false-belief tasks both impose demands beyond mental-state reasoning and capture only one aspect of ToM. \citet{apperly2012theory} further distinguishes understanding mental-state concepts from the cognitive processes supporting their deployment and the propensity to consider others' mental states. These distinctions suggest that answering a supplied belief question and spontaneously considering another agent's mental states during interaction should not be treated as interchangeable evidence.}

\revdel{LLM evaluations have broadened both the scope of mental-state reasoning and the evidence used to assess it. Hi-ToM examines higher-order beliefs and deceptive communication \citep{wu2023hitom}, while CogToM extends coverage across cognitive paradigms \citep{tong2026cogtom}. MeetingToM introduces multimodal evidence from recorded meetings to evaluate individual mental states, interpersonal attitudes, and group consensus \citep{wang2026meetingtom}. Beyond response accuracy, \citet{zhu2024beliefs} investigate belief-related information through decoding and interventions on model representations. NCP-ExploreToM evaluates whether models can plan actions that induce specified belief states rather than only answer questions about them \citep{slater2026planning}. These approaches move beyond a single task or outcome measure. However, interpreting provided situations or directing characters' actions differs from inferring the changing beliefs and strategies of an independently adapting opponent. This motivates examining what models can infer and use as evidence evolves through reciprocal interaction.}
\revadd{False-belief narratives test whether an LLM can answer what a character knows or believes \citep{kosinski2024evaluating,vanduijn2023theory}, but performance can be sensitive to small scenario changes \citep{ullman2023trivial}. Human ToM research cautions that these tasks also impose nonmental demands and sample only part of mentalizing \citep{bloom2000falsebelief}; knowing mental-state concepts need not imply their spontaneous use \citep{apperly2012theory}. LLM benchmarks now cover higher-order beliefs, broader cognitive paradigms, multimodal situations, and plans that alter another agent's beliefs \citep{wu2023hitom,tong2026cogtom,wang2026meetingtom,slater2026planning}. Representation analyses add another source of evidence \citep{zhu2024beliefs}. These approaches leave open how an LLM forecasts an independently adapting opponent during reciprocal play.}

\subsection{Strategic Interaction and Computational Opponent Modeling}

\revdel{Repeated strategic games couple prediction with action: players choose based on expectations about others, and observed choices provide evidence for subsequent adaptation. Such adaptation can draw on different sources, including behavioral regularities, experienced rewards, or representations of how an opponent learns and decides. Strategic success therefore does not, by itself, distinguish mental-state inference from other ways of predicting and responding to behavior.}

\revdel{Computational models make these alternatives more explicit. Recursive Bayesian $k$-ToM models describe agents that infer and simulate lower-level opponents' learning and decision processes, allowing competing accounts of behavior to be examined in repeated games \citep{devaine2014evolution,devaine2014social}. The \texttt{tomsup} framework makes these recursive agents and simpler alternatives available as both computational models and experimental opponents \citep{waade2023tomsup}. Rather than prescribing a recursive inference algorithm, ToMnet learns representations of agents' characteristics and current states from behavioral trajectories to predict subsequent behavior \citep{rabinowitz2018machine}. These approaches provide explicit hypotheses or learned models linking observations to behavior, although predictive success does not uniquely identify the representations or computations used by another system producing similar behavior.}

\revdel{Recent work brings strategic interaction and computational modeling to LLM evaluation. \citet{sohail2026mentalization} examine repeated inspection and rock--paper--scissors games against algorithmic opponents, combining behavioral outcomes with cognitive-model fitting to characterize mentalizing strategies. This provides evidence beyond aggregate success, although inferred processes remain conditional on the candidate models used to explain observed choices. Direct forecast evaluation and comparisons between supplied and withheld opponent information provide complementary evidence about what supports successful behavior. Our two-action hide-and-seek task adopts a minimal interactive setting in which explicitly specified opponents make these comparisons possible, while retaining opposing objectives and reciprocal adaptation.}
\revadd{Economic games have been used to study LLM social behavior. Trust games examine trust and reciprocity, including across multiple rounds \citep{ou2025social}; public-goods games examine contributions, cooperation, and collective returns \citep{li2025spontaneous}. Repeated matrix games test cooperation and coordination, and prompting a model to predict its partner's move can improve coordination \citep{akata2025playing}. These studies include repeated play, but choices and payoffs remain their principal endpoints. Even a next-action prediction does not by itself show whether the model tracks an opponent's changing state.}

\revadd{Recursive Bayesian $k$-ToM models describe agents that infer how others learn and choose \citep{devaine2014evolution,devaine2014social}; \texttt{tomsup} implements these agents alongside simpler strategies \citep{waade2023tomsup}. LLM studies have also fitted cognitive models to repeated-game choices \citep{sohail2026mentalization}. Such fits compare candidate accounts of behavior but need not identify the LLM's own computation. We use known opponent update rules to score forecasts and selected belief reports directly, separately from live payoff.}

\subsection{Improving ToM Reasoning and Its Deployment}

\revdel{Efforts to improve ToM address both the organization of inference and the learning of task-relevant behavior \citep{chen2025assessment}. Inference-time methods make intermediate representations more explicit. For example, RECTOM recursively constructs character perspectives to answer nested belief questions \citep{lei2026perspective}. Its gains demonstrate the usefulness of structured perspective construction on the evaluated narrative and dialogue benchmarks; whether those gains extend to reciprocal interaction with independently adapting opponents requires separate evaluation.}

\revdel{Post-training methods seek to strengthen these abilities through supervised or reinforcement learning. \citet{zhong2026shortcuts} examine shortcut-prone ToM datasets and compare supervised fine-tuning with reinforcement fine-tuning that elicits reasoning before answering. Their findings show why training gains must be interpreted alongside data properties and generalization tests. The principal rewards target output format and answer correctness, however, rather than directly validating every intermediate inference. In an applied setting, RebuttalAgent trains a ToM--Strategy--Response framework for academic rebuttal generation using supervised and reinforcement learning \citep{he2026dancing}. Such results support the practical value of structured social reasoning, but improved responses do not directly establish accurate recovery of a recipient's internal state.}

\revdel{Taken together, these literatures motivate examining not only whether elicitation and training improve outcomes, but also which information and abilities support the improvement. We connect repeated interaction with direct opponent prediction, fixed-history comparisons, and selected information-based diagnostics. The aim is not to identify a unique internal mechanism or require exact replication of a cognitive model, but to test how far behavioral and predictive gains support an interpretation in terms of inferring and using an opponent's internal states.}
\revadd{Before LLMs, ToMnet learned predictive representations of agents from their behavioral trajectories \citep{rabinowitz2018machine}. In SymmToM, reinforcement-learning agents that modeled other agents' mental states outperformed agents without such a model, but remained below agents given the true states \citep{sclar2022symmetric}. The former tests prediction of another agent's behavior; the latter tests whether state modeling helps an acting agent.}

\revadd{LLM ToM improvement work uses perspective construction, supervised or reinforcement training, and task-specific social-response systems \citep{chen2025assessment,lei2026perspective,zhong2026shortcuts,he2026dancing}. Stronger answers or responses need not show recovery of an opponent's changing state. We therefore test elicitation and training in live play alongside direct forecasts and fixed-history information controls.}

\section{Task and evaluation framework}
\label{sec:problem-setup}

Each \revnew{main} live condition pairs focal player $A$ with opponent $B$ for 20 matches of 50 simultaneous binary decisions. The seeker receives $+1$ for matching the opponent's action and $-1$ otherwise; the hider receives the opposite payoff (Figure~\ref{fig:task-schematic}C). History $h_t$ contains only completed rounds. We distinguish opponent action probability $q_t=P_B(a_t^B=1)$, opponent belief about the focal player $b_t=P_B(a_t^A=1)$, and full simulator state $z_t$: reference targets, not assumed LLM representations. ToM0 learns an action tendency; ToM1 and ToM2 model lower-order learners. Biased random (RB), Q-learning (QL), and win-stay/lose-shift (WSLS) are behavioral controls (Appendix~\ref{app:ktom}).

Natural play supplies instructions and history without opponent identity. Fixed-history studies vary information or response format while holding observations constant. ``Opponent specification'' supplies type, parameters, state schema and initialization, not current state or update equations. Exact-rule and true-state conditions are separately labeled. Reference best responses and actual action interfaces are distinguished in Appendix~\ref{app:ktom}.

Payoff sums 50 round rewards. Mean squared error (MSE) scores both the forecast of simulator $q_t$ and the report of simulator $b_t$; binary Brier score compares an action forecast with the opponent's sampled action~\citep{brier1950verification}. Constant $0.5$ and past-action frequency are prediction controls scored on the same decisions. Where shown, 95\% confidence intervals (CIs) are obtained by resampling complete matches, keeping the decisions within each match together. Separately collected live conditions are compared as independent samples. These CIs describe variation across evaluated matches, not across training seeds; no interval is implied for points without one. Appendix~\ref{app:metrics} gives the scoring protocol. New diagnostics use development data, not sealed-final evaluation.

\begin{figure}[!t]
\centering
\includegraphics[width=\linewidth]{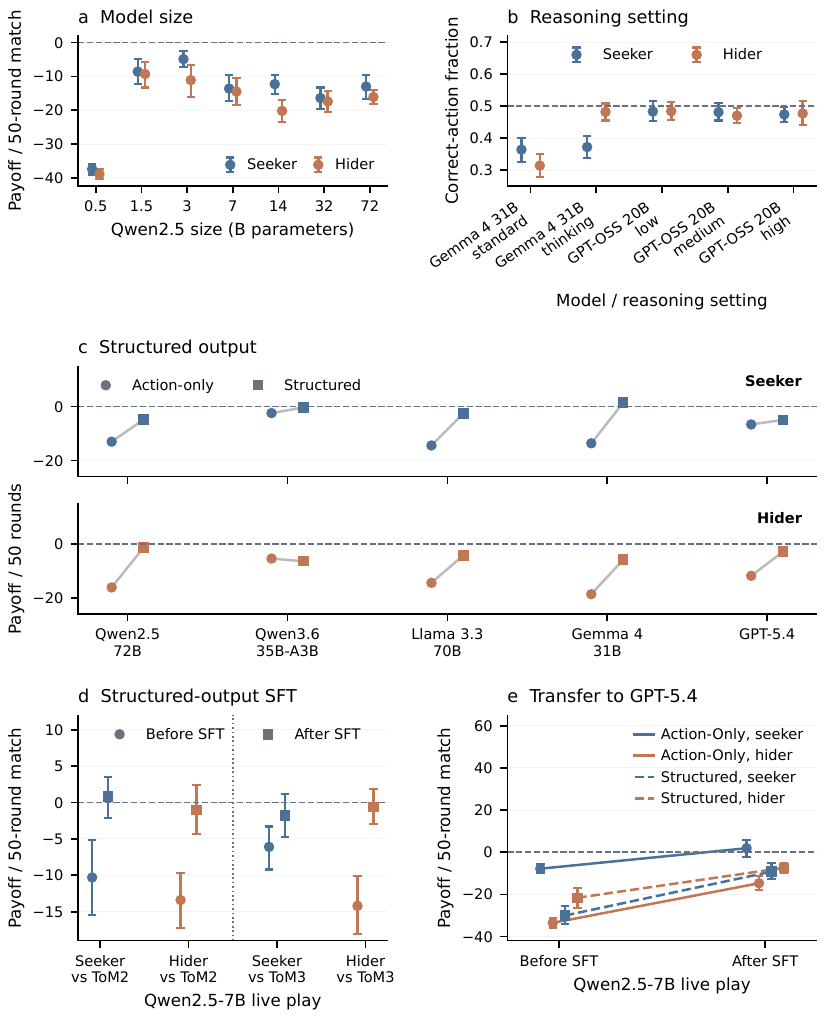}
\caption{\textbf{Live play across conditions.} Blue denotes seeker and orange hider, including in (c); circles/squares distinguish action-only/structured output, with line style also used in (e). In (d), circles/squares indicate before/after SFT. (a) Qwen2.5 size and (b) reasoning settings against ToM2. Gemma standard hider uses an audited historical action-only run; Gemma thinking and GPT-OSS hider settings use audited replacements, not paired controls. (c) Output-format comparisons by role. (d) Qwen2.5-7B structured-output behavioral supervised fine-tuning (SFT) against ToM2 and held-out ToM3; (e) action-only and structured-output behavioral SFT transfer against GPT-5.4. Each completed cell has 20 live matches of 50 rounds. Whiskers show 95\% match-bootstrap CIs where available; panel c omits intervals because archived seeker CIs cannot all be verified. Prespecified random fallback occurred in 30 Gemma thinking rounds and 3/4/10 GPT-OSS low/medium/high hider rounds; all rounds remain in the results. Missing cells are not zero.}
\label{fig:behavior}
\end{figure}

\section{Results}
\label{sec:results}

\subsection{Live-play gains depend on opponent, role and intervention}
\label{sec:diagnosis}

Figure~\ref{fig:behavior} compares model size, requested reasoning effort, structured output and training against ToM2, the highest prespecified opponent level, which models another agent's learning. Figure~\ref{fig:s1} presents live-play comparisons across both roles and six opponent classes.

\paragraph{Larger models and higher requested effort do not improve ToM2 payoff monotonically.}
Against ToM2, Qwen2.5 payoff is non-monotonic: seeker/hider scores are $-4.9/-11.1$ at 3B and $-13.0/-16.1$ at 72B (Figure~\ref{fig:behavior}a). Seeker effort settings show no consistent gain. The audited historical Gemma 4 standard hider cell has a $0.314$ correct-action fraction; the repaired thinking hider cell yields $0.482$ and $-1.8$ payoff per match (95\% CI $[-4.5,0.8]$), with 30 of 1,000 decisions using prespecified random fallback after final-channel truncation (Figure~\ref{fig:behavior}b). In 20-match GPT-OSS-20B hider evaluations, correct-action fractions are $0.484$, $0.470$, and $0.477$ at low, medium, and high effort, respectively; medium-effort payoff is $-3.0$ per match (95\% CI $[-5.4,-0.5]$). The completed six-opponent comparison also shows no uniform ordering of effort settings (Figure~\ref{fig:s1}). These separate cells mix historical and new runtimes and do not show a monotonic effort gain or a clean model-only comparison.

\paragraph{Structured output helps GPT-5.4 hiders but hurts Qwen2.5 against WSLS.}
An action-only response contains just the final action. The structured interface also asks for an opponent strategy estimate, predicted next action and role-conditioned best action before the final action. Later fields can condition on earlier text, but that does not establish that a reported prediction caused the choice (Appendix~\ref{app:chain-consistency}). In Figure~\ref{fig:behavior}c, the Qwen3.6-35B-A3B structured-minus-action-only change against ToM2 is $+2.1$ (95\% CI $[-1.0,5.4]$) as seeker and $-1.0$ $[-6.9,5.2]$ as hider; neither interval excludes zero. Fresh matched GPT-5.4 changes are $+1.7$ $[-1.2,4.5]$ and $+9.0$ $[3.9,14.1]$, respectively. These are separate role comparisons, not a role-interaction test. Across other opponents, audited Qwen2.5-32B WSLS changes are $-26.8$ $[-30.4,-22.9]$ as seeker and $-31.9$ $[-34.0,-29.6]$ as hider (Figure~\ref{fig:s2}). Historical open-model contrasts are unpaired and retain prompt, runtime and budget differences, including an ambiguous structured hider instruction (Appendix~\ref{app:prompts}).

\paragraph{Behavioral training improves familiar play but transfer remains role-sensitive.}
Behavioral SFT uses archived histories and role-conditioned action labels. Its action-only format supervises the final action; its structured-output format also supervises strategy and action-chain fields. Neither trains directly on continuous $b_t$ or $q_t$ (Appendix~\ref{app:sft}). We call these action-only and structured-output behavioral SFT, distinct from the later SFT studies that directly supervise $q_t$. Figures~\ref{fig:behavior}d,e and~\ref{fig:s3} evaluate new live matches, not replayed training histories. Against the training opponent ToM2, action-only behavioral SFT changes Qwen2.5-7B payoff from $-13.5$ to $+2.3$ as seeker and from $-14.4$ to $-1.2$ as hider (Figure~\ref{fig:s3}); structured-output behavioral SFT also improves both roles (Figure~\ref{fig:behavior}d). ToM3 was absent from training, so it tests higher-order within-game transfer. Across all six training opponents, some cells are weak or adverse (Figure~\ref{fig:s3}). Against GPT-5.4 as a distinct live opponent, the action-only endpoint earns $+1.9$ as seeker but $-14.7$ as hider (Figure~\ref{fig:behavior}e). The two SFT formats have unequal update budgets. These are behavioral gains, not direct evidence of belief recovery.

\revnew{\paragraph{Action-chain RL earns high WSLS payoff but low GPT-5.4 payoff.}
Archived reinforcement learning (RL) using group relative policy optimization (GRPO) extends the behavioral picture (Appendix~\ref{app:rl}). In a 600-update offline run, the final checkpoint earns $+38.3$/$+33.4$ against WSLS but $+1.0$/$+0.5$ against ToM2 and $-8.6$/$-17.5$ against GPT-5.4 (seeker/hider, 20 matches per cell). The archive lacks a matched pre-RL arm. Separate policy-refresh and probability-reward variants have mixed outcomes rather than a consistent gain with further updates. A fixed-history test of another recursive-GRPO checkpoint detects no clear forecast advantage for true over shuffled belief prefixes, although individual actions change (Appendix~\ref{app:rl-prefix}).}

\begin{figure}[!t]
\centering
\includegraphics[width=\linewidth]{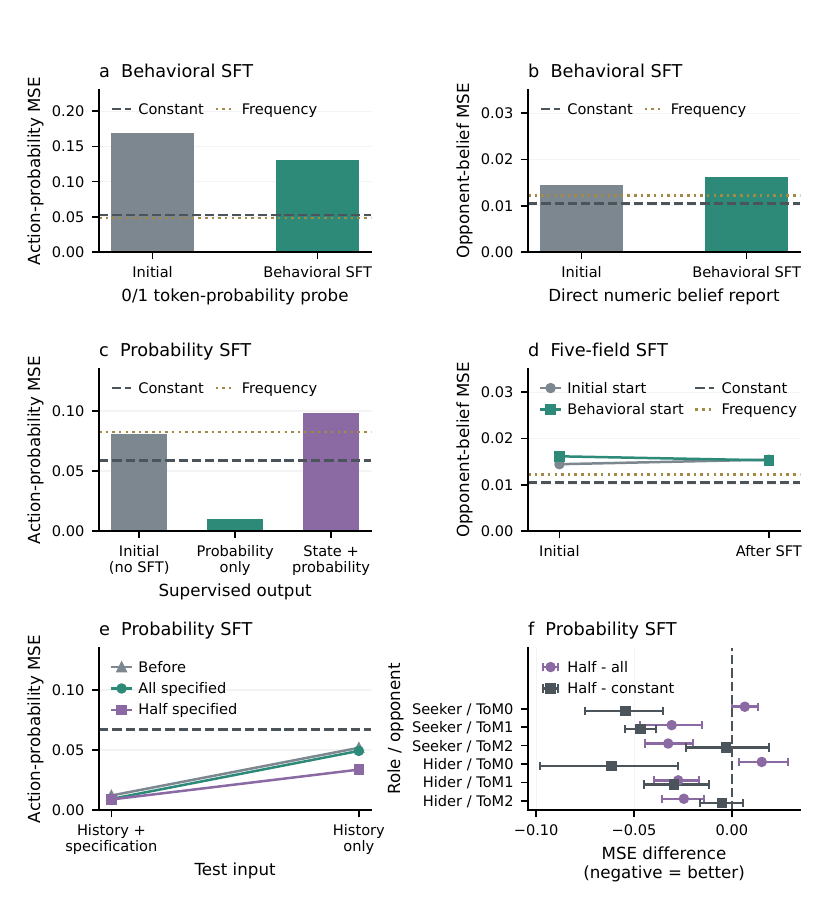}
\caption{\textbf{Prediction learning and belief reporting are distinct.} Qwen2.5-7B fixed-history studies. Colors distinguish training conditions, not roles: both roles are pooled except in (f), which separates them by row. (a,b) Action confidence and belief reports on six matches. (c) SFT on $q_t$ alone versus a numeric state vector and $q_t$, with opponent specification, 60 matches; invalid outputs receive loss 1. (d) Five-field value SFT, including belief targets, on six matches. ``Initial'' marks the start of each five-field run; one run starts from the initial model and the other from a behavioral-SFT checkpoint. (e,f) Equal-update probability-only SFT continuations on a fresh R5 cohort, with specification on all versus half of training examples; (e) action-probability MSE by test input and (f) history-only MSE differences by opponent. Both use 60 fresh matches and 540 noninitial predictions per condition. Dashed lines mark constant baselines; (f) shows 95\% paired match-bootstrap intervals. Panels use distinct cohorts and interfaces (Appendix~\ref{app:qwen}).}
\label{fig:qwen}
\end{figure}

\subsection{Action-probability supervision improves fixed-history forecasts}
\label{sec:learning}

The live-payoff gains in Figure~\ref{fig:behavior} do not tell us whether a model learned the opponent's action probability or belief. Figure~\ref{fig:qwen} tests these outputs on fixed histories, where predictions cannot change the next observation.

\paragraph{Behavioral gains do not carry over to accurate belief reports.}
On six shared ten-round histories, structured-output behavioral-SFT checkpoint action-probability MSE is $0.13126$, versus initial $0.16828$ and constant $0.05230$ (Figure~\ref{fig:qwen}a). This is normalized binary-token confidence, not decimal elicitation. Belief MSE is $0.01614$ versus constant $0.01052$ (b). A separate five-field value SFT targets belief, action probability, opponent action, best action and final action. Its selected belief MSEs are $0.01542$ and $0.01527$ from the initial and behavioral starts, respectively (d), still above constant. These six-match probes use interfaces that differ from live play (Appendix~\ref{app:qwen}).

\paragraph{Probability-only SFT reduces forecast error with supplied specification.}
\revold{Separate LoRA~\citep{hu2021lora} predictors learn action probability alone or full state plus probability. Both receive history and specification. On 60 held-out development matches, probability-only SFT reduces MSE from $0.08119$ to $0.00990$, below constant $0.05874$ (Figure~\ref{fig:qwen}c). The state package has penalized MSE $0.09814$, with 582/600 valid predictions and valid-only MSE $0.07024$. Its exposure repair was rejected by validation. Different output burdens, warmup information and recurrence preclude a pure belief-utility ablation.}
\revnew{Separate low-rank adaptation (LoRA)~\citep{hu2021lora} predictors receive history and opponent specification. Probability-only SFT trains on action probability $q_t$ alone; state-plus-probability SFT trains on a numeric state vector and $q_t$. On 60 held-out development matches, probability-only SFT reduces MSE from $0.08119$ to $0.00990$, below constant $0.05874$ (Figure~\ref{fig:qwen}c). State-plus-probability SFT has penalized MSE $0.09814$, with 97.0\% of the 600 predictions valid and valid-only MSE $0.07024$. Its exposure repair was rejected by validation. The arms also differ in output burden, warmup information and recurrence, so this is not a pure belief-utility ablation.}

\paragraph{Mixed exposure reduces history-only forecast error.}
On the fresh R5 cohort plotted in Figure~\ref{fig:qwen}e, removing specification increases the starting checkpoint's MSE from $0.01206$ to $0.05182$ on the same 540 noninitial decisions. We compare two 200-update continuations of probability-only SFT with identical targets and row order: a specification-only continuation, whose training examples all include the bundle, and a mixed-exposure continuation, whose examples omit it half the time. On 60 fresh history-only development matches, MSE falls from $0.04926$ to $0.03367$, a 31.65\% reduction (paired difference $-0.01559$, 95\% CI $[-0.02054,-0.01071]$; Figure~\ref{fig:qwen}e,f). ToM1/2 improve in both roles; ToM0 worsens. ToM2 point estimates beat constant, but both intervals cross zero. The gain is confined to fixed-history action forecasts.

\begin{figure}[!t]
\centering
\includegraphics[width=\linewidth]{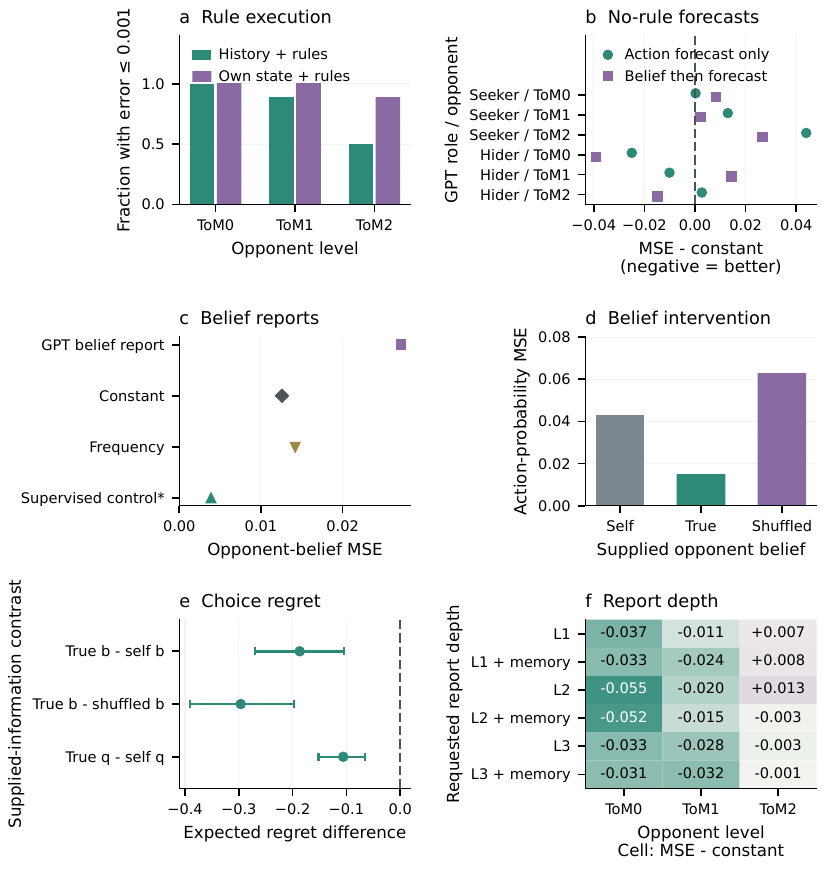}
\caption{\textbf{Using supplied opponent information differs from inferring it.} GPT-5.4 medium on fixed histories. Colors encode conditions, not roles; only (b) separates them. (a) Known-rule execution, six matches; incomplete chains count as errors. (b,c) No-rule replication: 24 matches, 216 noninitial points per arm; the supervised control has additional task-specific training. In the six-match supplied-belief intervention, (d) plots forecast MSE and (e) plots expected-regret differences across 54 decisions per arm. The 66.7\% to 96.3\% reference-best-action agreement is a separate audit summary of these decisions (Appendix~\ref{app:gpt}), not a plotted axis. Regret whiskers show paired 95\% match-bootstrap CIs. (f) Six-match report-depth pilot: forecast MSE minus constant, without intervals. These studies are separate; Appendix~\ref{app:gpt} details retry-assisted coverage.}
\label{fig:gpt}
\end{figure}

\subsection{Supplied beliefs guide decisions more reliably than autonomous reports}
\label{sec:gpt}

Qwen's forecasts depend on what is supplied and what the model has seen during training. We use GPT-5.4 to separate three related abilities: following known rules, inferring beliefs without those rules, and using a correct belief supplied at decision time (Figure~\ref{fig:gpt}).

\paragraph{Known-rule execution is a positive control.}
\revold{GPT-5.4 medium executes supplied recursive rules using its own maintained full state: action probabilities are within $0.001$ on 52/54 scheduled noninitial points, and five of six chains finish (Figure~\ref{fig:gpt}a). The unfinished chain is the seeker against ToM2; its missing decisions remain in the denominator. This is substantial known-rule execution, not autonomous opponent identification.}
\revnew{With supplied recursive rules and self-generated prior states, GPT-5.4 at medium reasoning effort produces action probabilities within $0.001$ on 96.3\% of the 54 scheduled noninitial points, and five of six chains finish (Figure~\ref{fig:gpt}a). The unfinished seeker-ToM2 chain contributes its missing decisions to the denominator. State-coordinate accuracy is weaker (Appendix~\ref{app:gpt}).}

\paragraph{Added belief reports do not reliably improve inference.}
Without rules or identity, a separate 24-match replication compares a one-field $q$ report (L1) with a two-field $b,q$ report (L2). Forecast MSE is $0.06364$ versus $0.05910$; the 95\% paired CI for the difference is $[-0.01511,0.00823]$ (Figure~\ref{fig:gpt}b). L2 improves hider-ToM2 forecast MSE descriptively, but seeker-ToM2 remains above constant. L2 belief MSE $0.02717$ is worse than constant $0.01258$ (c). A trained same-observation predictor does better but has additional task-specific experience. The six-match depth-by-memory pilot (f) has mixed descriptive differences, too small to establish a general gain from nesting or retained outputs. Report depth is not opponent order.

\paragraph{Correct supplied beliefs improve downstream choices.}
\revold{Replacing self-reported $b$ with true $b$ on the same 54 decisions reduces forecast MSE from $0.04319$ to $0.01484$ and increases oracle-best choices from 36/54 to 52/54 (Figure~\ref{fig:gpt}d,e). Expected regret falls by $0.18676$ (95\% paired CI $[-0.26968,-0.10510]$), also improving over shuffled $b$. This is external information use, not identification of a neural representation; a simple direction rule may suffice. Fixed-history regret is not live payoff. Appendix~\ref{app:gpt} separates original failures and retry-assisted coverage.}
\revnew{Replacing self-reported $b$ with simulator $b_t$ on the same 54 decisions reduces forecast MSE from $0.04319$ to $0.01484$ (Figure~\ref{fig:gpt}d). In the same audited decisions, reference-best-action agreement, meaning agreement with the role-conditioned best action computed from simulator $q_t$, rises from 66.7\% to 96.3\% (Appendix~\ref{app:gpt}). Expected regret falls by $0.18676$ (95\% paired CI $[-0.26968,-0.10510]$; Figure~\ref{fig:gpt}e), also improving over shuffled $b$. A simple direction rule could use the supplied belief; this intervention does not identify an internal representation. Fixed-history regret is not live payoff. Appendix~\ref{app:gpt} separates original failures and retry-assisted coverage.}

\ifrevisionmarkup\clearpage\fi
\section{Discussion, limitations and conclusion}
\label{sec:discussion}

\revold{What separates learning to play better from learning to model an adaptive opponent? In this game, many probability estimates imply the same best response because only their position relative to $0.5$ determines the preferred action. Payoff therefore leaves much of the underlying inference unresolved. Our comparisons examine the additional requirement: constructing opponent information that supports predictions as behavior changes.\par
The supplied-information tests locate a gap between using information and inferring it. Qwen2.5-7B after probability-only SFT forecasts accurately with opponent specification, but removing that specification increases MSE more than fourfold. Separately, supplying true beliefs to GPT-5.4 raises best-response choices from 66.7\% to 96.3\%, although its autonomous belief reports remain less accurate than a constant baseline (Figures~\ref{fig:qwen} and~\ref{fig:gpt}). Successful downstream decisions can thus depend on information the model cannot reliably reconstruct. This makes inference from history a concrete target for learning.\par
Training exposure partly addresses that target. At a fixed update budget, removing opponent specification from half of the Qwen probability-SFT continuation examples reduces history-only MSE by 31.65\%. A model trained with supplied specification can learn a conditional prediction rule; mixed exposure also requires predictions without that support. Yet ToM1/2 improve while ToM0 deteriorates (Figure~\ref{fig:qwen}f). The aggregate gain therefore comes with a tradeoff across opponent dynamics, which matters for an agent that must adapt without knowing which opponent it faces.\par
Exposure also helps frame the difficulty encountered in the RL experiments. The 120-update probability-reward run on Qwen2.5-7B-Instruct achieves over 99\% consistency between output fields while forecast MSE remains above the constant baseline (Appendix~\ref{app:rl-probability}). It learns to produce mutually consistent forecasts and responses without achieving accurate forecasts on fresh histories. Because those updates reuse one collected buffer and evaluation generates histories under the updated policy, distribution shift is a candidate bottleneck. Rewarding a forecast more directly still requires learning across the observations that a changing policy encounters.\par
The main limitation is that these comparisons do not yet connect improved inference to sustained live adaptation. Fixed-history tests hold observations constant, whereas an agent's actions change its opponent and subsequent evidence. The next test should deploy the probability-only and mixed-exposure checkpoints through the same action rule and measure whether their forecast gains persist. Belief interventions should then test whether correct content improves those forecasts under matched input formats; the trained belief-prefix diagnostic currently leaves that benefit uncertain (Figure~\ref{fig:rl-prefix}). Replication across training seeds and larger intervention cohorts is needed to assess stability. Finally, simulator beliefs provide exact targets within specified algorithms; extending the account to human opponents requires independent evidence for the psychological relevance of those targets.\par
Together, these results show that learning to use opponent information can outpace learning to infer it. Mixed exposure narrows this gap, while the RL results show why response consistency alone is an inadequate training endpoint. The central criterion for adaptive opponent modeling is whether inferred information improves predictions on the histories an agent creates through its own actions.}

\revnew{\textbf{Better play does not establish that an agent models its opponent.} In this binary game, every forecast on the same side of $0.5$ yields the same best response, so payoff can improve while the probability estimate remains inaccurate. The supplied-information tests make this distinction concrete. Qwen2.5-7B after probability-only SFT forecasts accurately when given opponent specification, but removing it increases MSE more than fourfold (Figure~\ref{fig:qwen}e). Separately, supplying simulator beliefs raises GPT-5.4's reference-best-action agreement from 66.7\% to 96.3\% (Appendix~\ref{app:gpt}), while its autonomous belief reports remain less accurate than a constant baseline (Figure~\ref{fig:gpt}c). These are not matched model comparisons; both show a gap between using useful opponent information and inferring it from history.}

\revnew{\textbf{Training gains are uneven when opponent information must be inferred from history.} The mixed-exposure Qwen continuation omits opponent specification from half of the probability-only SFT examples while holding targets, row order and update count fixed. Its history-only MSE falls by 31.65\% relative to the specification-only continuation, with improvements for ToM1/2 in both roles but worse performance on ToM0 (Figure~\ref{fig:qwen}f). The RL results expose a different limit: a 120-update probability-reward run on Qwen2.5-7B-Instruct achieves over 99\% consistency between output fields, yet forecast MSE exceeds the constant baseline on fresh histories (Appendix~\ref{app:rl-probability}). Because the updates reused one collected buffer while evaluation generated histories under the updated policy, distribution shift is a candidate bottleneck, not an established cause. A coherent response still needs an accurate forecast as input.}

\revnew{\textbf{Using a belief field is different from using its correct content.} On GPT-5.4's fixed histories, supplying simulator beliefs rather than shuffled beliefs reduces expected regret by $0.29640$ (95\% CI $[0.19726,0.39106]$; Appendix~\ref{app:gpt}). In a separate recursive-GRPO prefix diagnostic, true and shuffled beliefs produce different final actions on 10.3\% of decisions, but the forecast-MSE advantage of true beliefs remains uncertain (Figure~\ref{fig:rl-prefix}). These experiments are not a matched comparison of models or training methods. Together they show why field sensitivity and a benefit specific to correct belief content must be tested separately.}

\revnew{\textbf{Limitations.} These results come from one two-action game, where distinct beliefs and reactive heuristics can yield the same choices and payoff. Future tasks should match simple action and reward cues across opponents but make belief-based and reactive strategies diverge after diagnostic feedback. Fixed-history forecasts also omit feedback generated by the agent's own actions. A matched live deployment of the probability-only and mixed-exposure checkpoints must test whether their forecast gains persist. Format-matched belief interventions, more training seeds and larger cohorts would test content-specific benefits and stability. Simulator states are exact targets for these algorithms, not validated human mental states.}

\revnew{\textbf{Conclusion.} In this game, better play is evidence of adaptation, not by itself of theory of mind. A stronger theory-of-mind interpretation would require accurate, content-sensitive predictions on live histories and tasks where belief-based and simpler strategies lead to different actions.}

\label{main-text-end}
\clearpage
\section*{Acknowledgments}
We thank Maxine Sherman and James Bennett for helpful discussions. This work was supported by research startup funding from the Georgia Institute of Technology to R.C.W. This research was supported in part through research cyberinfrastructure resources and services provided by the Partnership for an Advanced Computing Environment (PACE) at the Georgia Institute of Technology, Atlanta, Georgia, USA.

\section*{AI use statement}
Generative AI assisted with study design, code, data and artifact checks, analysis, figures, and manuscript drafting and editing. The authors reviewed and verified the reported methods, results, and claims and take responsibility for the paper.

\section*{Code and data availability}
Code, selected model outputs, evaluated adapters, figure data, and instructions for reanalysis are available at \url{https://github.com/xhb120633/opponent-modeling}. The repository documents which analyses can be reproduced from released data and which require separate model weights, API access, or compute.

\bibliography{references}

@article{premack1978chimpanzee,
  author  = {Premack, David and Woodruff, Guy},
  title   = {Does the Chimpanzee Have a Theory of Mind?},
  journal = {Behavioral and Brain Sciences},
  year    = {1978},
  volume  = {1},
  number  = {4},
  pages   = {515--526},
  doi     = {10.1017/S0140525X00076512}
}

@article{kosinski2024evaluating,
  author  = {Kosinski, Michal},
  title   = {Evaluating Large Language Models in Theory of Mind Tasks},
  journal = {Proceedings of the National Academy of Sciences},
  year    = {2024},
  volume  = {121},
  number  = {45},
  pages   = {e2405460121},
  doi     = {10.1073/pnas.2405460121}
}

@inproceedings{vanduijn2023theory,
  author    = {van Duijn, Max and van Dijk, Bram and Kouwenhoven, Tom and de Valk, Werner and Spruit, Marco and van der Putten, Peter},
  title     = {Theory of Mind in Large Language Models: Examining Performance of 11 State-of-the-Art Models vs. Children Aged 7--10 on Advanced Tests},
  booktitle = {Proceedings of the 27th Conference on Computational Natural Language Learning (CoNLL)},
  year      = {2023},
  pages     = {389--402},
  publisher = {Association for Computational Linguistics},
  doi       = {10.18653/v1/2023.conll-1.25},
  url       = {https://aclanthology.org/2023.conll-1.25/}
}

@article{ullman2023trivial,
  author        = {Ullman, Tomer},
  title         = {Large Language Models Fail on Trivial Alterations to Theory-of-Mind Tasks},
  journal       = {arXiv preprint arXiv:2302.08399},
  year          = {2023},
  eprint        = {2302.08399},
  archiveprefix = {arXiv},
  primaryclass  = {cs.AI},
  doi           = {10.48550/arXiv.2302.08399},
  url           = {https://arxiv.org/abs/2302.08399}
}

@inproceedings{zhu2024beliefs,
  author    = {Zhu, Wentao and Zhang, Zhining and Wang, Yizhou},
  title     = {Language Models Represent Beliefs of Self and Others},
  booktitle = {Proceedings of the 41st International Conference on Machine Learning},
  year      = {2024},
  volume    = {235},
  series    = {Proceedings of Machine Learning Research},
  pages     = {62638--62681},
  publisher = {PMLR},
  url       = {https://proceedings.mlr.press/v235/zhu24o.html}
}

@article{devaine2014evolution,
  author  = {Devaine, Marie and Hollard, Guillaume and Daunizeau, Jean},
  title   = {Theory of Mind: Did Evolution Fool Us?},
  journal = {PLOS ONE},
  year    = {2014},
  volume  = {9},
  number  = {2},
  pages   = {e87619},
  doi     = {10.1371/journal.pone.0087619}
}

@article{devaine2014social,
  author  = {Devaine, Marie and Hollard, Guillaume and Daunizeau, Jean},
  title   = {The Social Bayesian Brain: Does Mentalizing Make a Difference When We Learn?},
  journal = {PLOS Computational Biology},
  year    = {2014},
  volume  = {10},
  number  = {12},
  pages   = {e1003992},
  doi     = {10.1371/journal.pcbi.1003992}
}

@article{waade2023tomsup,
  author  = {Waade, Peter T. and Enevoldsen, Kenneth C. and Vermillet, Arnault-Quentin and Simonsen, Arndis and Fusaroli, Riccardo},
  title   = {Introducing {tomsup}: Theory of Mind Simulations Using {Python}},
  journal = {Behavior Research Methods},
  year    = {2023},
  volume  = {55},
  number  = {5},
  pages   = {2197--2231},
  doi     = {10.3758/s13428-022-01827-2}
}

@inproceedings{wu2023hitom,
  author    = {Wu, Yufan and He, Yinghui and Jia, Yilin and Mihalcea, Rada and Chen, Yulong and Deng, Naihao},
  title     = {{Hi-ToM}: A Benchmark for Evaluating Higher-Order Theory of Mind Reasoning in Large Language Models},
  booktitle = {Findings of the Association for Computational Linguistics: EMNLP 2023},
  year      = {2023},
  pages     = {10691--10706},
  publisher = {Association for Computational Linguistics},
  doi       = {10.18653/v1/2023.findings-emnlp.717},
  url       = {https://aclanthology.org/2023.findings-emnlp.717/}
}

@inproceedings{tong2026cogtom,
  author    = {Tong, Haibo and Yue, Zeyang and Zhao, Feifei and Lin, Erliang and Jia, Lu and Chen, Ruolin and Sun, Yinqian and Zhang, Qian and Zeng, Yi},
  title     = {{CogToM}: A Comprehensive Theory of Mind Benchmark Inspired by Human Cognition for Large Language Models},
  booktitle = {Proceedings of the 64th Annual Meeting of the Association for Computational Linguistics (Volume 1: Long Papers)},
  year      = {2026},
  pages     = {31383--31425},
  publisher = {Association for Computational Linguistics},
  doi       = {10.18653/v1/2026.acl-long.1448},
  url       = {https://aclanthology.org/2026.acl-long.1448/}
}

@article{bloom2000falsebelief,
  author  = {Bloom, Paul and German, Tim P.},
  title   = {Two Reasons to Abandon the False Belief Task as a Test of Theory of Mind},
  journal = {Cognition},
  year    = {2000},
  volume  = {77},
  number  = {1},
  pages   = {B25--B31},
  doi     = {10.1016/S0010-0277(00)00096-2}
}

@article{apperly2012theory,
  author  = {Apperly, Ian A.},
  title   = {What Is ``Theory of Mind''? Concepts, Cognitive Processes and Individual Differences},
  journal = {Quarterly Journal of Experimental Psychology},
  year    = {2012},
  volume  = {65},
  number  = {5},
  pages   = {825--839},
  doi     = {10.1080/17470218.2012.676055}
}

@inproceedings{rabinowitz2018machine,
  author    = {Rabinowitz, Neil and Perbet, Frank and Song, Francis and Zhang, Chiyuan and Eslami, S. M. Ali and Botvinick, Matthew},
  title     = {Machine Theory of Mind},
  booktitle = {Proceedings of the 35th International Conference on Machine Learning},
  year      = {2018},
  volume    = {80},
  series    = {Proceedings of Machine Learning Research},
  pages     = {4218--4227},
  publisher = {PMLR},
  url       = {https://proceedings.mlr.press/v80/rabinowitz18a.html}
}

@inproceedings{chen2025assessment,
  author    = {Chen, Ruirui and Jiang, Weifeng and Qin, Chengwei and Tan, Cheston},
  title     = {Theory of Mind in Large Language Models: Assessment and Enhancement},
  booktitle = {Proceedings of the 63rd Annual Meeting of the Association for Computational Linguistics (Volume 1: Long Papers)},
  year      = {2025},
  pages     = {31539--31558},
  publisher = {Association for Computational Linguistics},
  doi       = {10.18653/v1/2025.acl-long.1522},
  url       = {https://aclanthology.org/2025.acl-long.1522/}
}

@article{lei2026perspective,
  author        = {Lei, Chao and Hu, Guang and Yang, Meng and Jiang, Yanbei and Lipovetzky, Nir},
  title         = {Mind the Perspective: Let's Reason Recursively for Theory of Mind},
  journal       = {arXiv preprint arXiv:2606.11724},
  year          = {2026},
  eprint        = {2606.11724},
  archiveprefix = {arXiv},
  doi           = {10.48550/arXiv.2606.11724},
  url           = {https://arxiv.org/abs/2606.11724}
}

@article{zhong2026shortcuts,
  author        = {Zhong, Jike and Lai, Yuxiang and Li, Ming and Li, Yuheng and Liu, Wuao and Dariush, Behzad and Psounis, Konstantinos and Lo, Shao-Yuan},
  title         = {From Shortcuts to Reasoning: Robust Post-Training of Theory of Mind with Reinforcement Learning},
  journal       = {arXiv preprint arXiv:2606.09092},
  year          = {2026},
  eprint        = {2606.09092},
  archiveprefix = {arXiv},
  primaryclass  = {cs.LG},
  doi           = {10.48550/arXiv.2606.09092},
  url           = {https://arxiv.org/abs/2606.09092}
}

@article{wang2026meetingtom,
  author        = {Wang, Ziyi and Wu, Yuhang and Piao, Dongxu and Liu, Xingyu and Zhou, Tianhui and Liu, Miao},
  title         = {{MeetingToM}: Evaluating Multimodal {LLMs} on Theory-of-Mind Reasoning in Multi-Party Meetings},
  journal       = {arXiv preprint arXiv:2607.19235},
  year          = {2026},
  eprint        = {2607.19235},
  archiveprefix = {arXiv},
  primaryclass  = {cs.CL},
  doi           = {10.48550/arXiv.2607.19235},
  url           = {https://arxiv.org/abs/2607.19235}
}

@article{slater2026planning,
  author        = {Slater, Ben and Mecattaf, Matteo G. and Cheke, Lucy G. and Burden, John and Street, Winnie},
  title         = {Theory of Mind and Persuasion Beyond Conversation: Assessing the Capacity of {LLMs} to Induce Belief States via Planning and Action},
  journal       = {arXiv preprint arXiv:2606.31916},
  year          = {2026},
  eprint        = {2606.31916},
  archiveprefix = {arXiv},
  primaryclass  = {cs.CL},
  doi           = {10.48550/arXiv.2606.31916},
  url           = {https://arxiv.org/abs/2606.31916}
}

@article{sohail2026mentalization,
  author        = {Sohail, Aamir and Zhong, Xintong and Konovalov, Arkady and Lockwood, Patricia L. and Zhang, Lei},
  title         = {Assessing Mentalization in Humans and Large Language Models},
  journal       = {arXiv preprint arXiv:2608.26291},
  year          = {2026},
  eprint        = {2608.26291},
  archiveprefix = {arXiv},
  primaryclass  = {cs.AI},
  doi           = {10.48550/arXiv.2608.26291},
  url           = {https://arxiv.org/abs/2608.26291}
}

@inproceedings{he2026dancing,
  author    = {He, Zhitao and Lyu, Zongwei and Fung, Yi R.},
  title     = {Dancing in Chains: Strategic Persuasion in Academic Rebuttal via Theory of Mind},
  booktitle = {Proceedings of the 14th International Conference on Learning Representations},
  year      = {2026},
  url       = {https://openreview.net/forum?id=zkCZDdeS9s}
}

@article{brier1950verification,
  author = {Brier, Glenn W.},
  title = {Verification of Forecasts Expressed in Terms of Probability},
  journal = {Monthly Weather Review},
  year = {1950},
  volume = {78},
  number = {1},
  pages = {1--3},
  doi = {10.1175/1520-0493(1950)078<0001:VOFEIT>2.0.CO;2},
  url = {https://journals.ametsoc.org/view/journals/mwre/78/1/1520-0493_1950_078_0001_vofeit_2_0_co_2.xml}
}

@article{hu2021lora,
  author = {Hu, Edward J. and Shen, Yelong and Wallis, Phillip and Allen-Zhu, Zeyuan and Li, Yuanzhi and Wang, Shean and Wang, Lu and Chen, Weizhu},
  title = {{LoRA}: Low-Rank Adaptation of Large Language Models},
  journal = {arXiv preprint arXiv:2106.09685},
  year = {2021},
  eprint = {2106.09685},
  archiveprefix = {arXiv},
  doi = {10.48550/arXiv.2106.09685},
  url = {https://arxiv.org/abs/2106.09685}
}

@article{shao2024deepseekmath,
  author = {Shao, Zhihong and Wang, Peiyi and Zhu, Qihao and Xu, Runxin and Song, Junxiao and Bi, Xiao and Zhang, Haowei and Zhang, Mingchuan and Li, Y. K. and Wu, Y. and Guo, Daya},
  title = {{DeepSeekMath}: Pushing the Limits of Mathematical Reasoning in Open Language Models},
  journal = {arXiv preprint arXiv:2402.03300},
  year = {2024},
  eprint = {2402.03300},
  archiveprefix = {arXiv},
  doi = {10.48550/arXiv.2402.03300},
  url = {https://arxiv.org/abs/2402.03300}
}

@inproceedings{sclar2022symmetric,
  author    = {Sclar, Melanie and Neubig, Graham and Bisk, Yonatan},
  title     = {Symmetric Machine Theory of Mind},
  booktitle = {Proceedings of the 39th International Conference on Machine Learning},
  series    = {Proceedings of Machine Learning Research},
  volume    = {162},
  pages     = {19450--19466},
  year      = {2022},
  publisher = {PMLR},
  url       = {https://proceedings.mlr.press/v162/sclar22a.html}
}

@article{ou2025social,
  author        = {Ou, Jiamin and Eikmans, Emile and Buskens, Vincent and Pankowska, Paulina and Shan, Yuli},
  title         = {Social Preferences with Unstable Interactive Reasoning: Large Language Models in Economic Trust Games},
  journal       = {arXiv preprint arXiv:2505.17053},
  year          = {2025},
  eprint        = {2505.17053},
  archiveprefix = {arXiv},
  url           = {https://arxiv.org/abs/2505.17053}
}

@inproceedings{li2025spontaneous,
  author    = {Li, Yuxuan and Shirado, Hirokazu},
  title     = {Spontaneous Giving and Calculated Greed in Language Models},
  booktitle = {Proceedings of the 2025 Conference on Empirical Methods in Natural Language Processing},
  pages     = {5271--5286},
  year      = {2025},
  publisher = {Association for Computational Linguistics},
  doi       = {10.18653/v1/2025.emnlp-main.267},
  url       = {https://aclanthology.org/2025.emnlp-main.267/}
}

@article{akata2025playing,
  author  = {Akata, Elif and Schulz, Lion and Coda-Forno, Julian and Oh, Seong Joon and Bethge, Matthias and Schulz, Eric},
  title   = {Playing Repeated Games with Large Language Models},
  journal = {Nature Human Behaviour},
  volume  = {9},
  pages   = {1380--1390},
  year    = {2025},
  doi     = {10.1038/s41562-025-02172-y},
  url     = {https://www.nature.com/articles/s41562-025-02172-y}
}
\bibliographystyle{author_year_refs}
\clearpage
\appendix
\renewcommand{\thefigure}{S\arabic{figure}}
\renewcommand{\theHfigure}{S\arabic{figure}}
\setcounter{figure}{0}
\section{Game, opponents and simulator targets}
\label{app:ktom}

\subsection{Decision-time information and payoffs}
Let $a_t^A,a_t^B\in\{0,1\}$ denote simultaneous actions. For a focal seeker,
\begin{equation}
u_{\rm S}(a_t^A,a_t^B)=2\mathbb{1}[a_t^A=a_t^B]-1,\qquad
u_{\rm H}(a_t^A,a_t^B)=-u_{\rm S}(a_t^A,a_t^B).
\end{equation}
The history is $h_t=\{(a_\tau^A,a_\tau^B,r_\tau^A)\}_{\tau<t}$. Role and public payoff rules are available before the first decision. Current and future actions are not prompt inputs. In live evaluation, the LLM chooses its action and therefore affects both the opponent's subsequent updates and future observations. In fixed-history evaluation, the focal actions were generated or recorded previously: changing a forecast does not alter the trajectory.

For a forecast $\hat q_t$, reference best responses are
\begin{equation}
\operatorname{BR}_{\rm S}(\hat q_t)=\mathbb{1}[\hat q_t>1/2],\qquad
\operatorname{BR}_{\rm H}(\hat q_t)=1-\operatorname{BR}_{\rm S}(\hat q_t).
\label{eq:role-response}
\end{equation}
Either action is optimal at a tie. This is a scoring reference, not an external decision rule used by every LLM condition.

For scoring, $q_t$ is the simulator's probability of action 1 before sampling or observing round $t$, and $b_t$ is its prediction that the focal player chooses 1. In the archived binary implementation these quantities are named \texttt{p\_self} and \texttt{p\_op}, respectively, from the algorithmic opponent's perspective. Names are not sufficient provenance: an oracle-export file can replace a prediction column. Audits therefore retain original model responses and join only necessary target columns by unique match/round keys.

\subsection{Recursive agents and state sufficiency}
The $k$-ToM opponents use the recursive Bayesian learning and decision procedures of \citet{devaine2014evolution,devaine2014social}, implemented in \texttt{tomsup}~\citep{waade2023tomsup}. ToM0 learns an opponent's action tendency; higher orders consider lower-order agents whose own predictions and learning depend on interaction. Schematically, the implemented computations are
\begin{equation}
z_{t+1}=F_{k,\theta}(z_t,a_t^B,a_t^A),\quad
b_t=H_{k,\theta}(z_t),\quad
q_t=G_{k,\theta,\operatorname{role}(B)}(z_t).
\label{eq:simulator}
\end{equation}
These symbols refer to the actual simulator functions, not a new inference algorithm proposed here. Higher-order states include estimates of lower-order agents and mixture probabilities over candidate orders. State coordinates include means in logit space, log variances and mixture weights; they should not be averaged into a single error without regard to units. The full-state training package removes cached readout probabilities and serializes 2, 13 and 37 numeric coordinates for ToM0, ToM1 and ToM2. Coordinate names, ordering and public initial values are included in the frozen data manifest and prompts.

The controlled Qwen probability-training family fixes \texttt{volatility=-2}, \texttt{b\_temp=-1}, and \texttt{bias=0}. These are the implementation's parameter values, not newly interpreted physical units. Exact update equations or executable source are supplied only in designated GPT rule conditions. Supplying the specification in the Qwen experiments gives identity, parameters and state schema, but no exact update procedure or current state.

A scalar $b_t$ or $q_t$ need not be sufficient to update a higher-order agent. A local program-reference check constructed states with the same displayed probability but different next updates; replay with the true rule and initialization reconstructed the reference beliefs to numerical tolerance. This distinguishes insufficient scalar summaries from a proof that inference from history is impossible. Unknown identities, parameters and finite histories can leave multiple state explanations compatible with observed actions. No claim of general identifiability follows from an accurate supervised predictor.

\subsection{Control opponents and transfer}
RB samples with a fixed action bias; QL adapts action values from rewards; WSLS repeats or switches according to the preceding outcome. The benchmark uses their archived \texttt{tomsup} configurations, rather than selecting parameters retrospectively to favor an LLM. The principal controlled cells cross both focal roles with ToM0, ToM1, and ToM2. ToM3--5 are within-game higher-order transfer conditions for the historical training series. GPT as a live opponent is a separate source of adaptive behavior, not interchangeable with GPT as the predictor in Figure~\ref{fig:gpt}. None of these transfer settings is a newly sampled human opponent population.

\section{Metrics, controls and statistical units}
\label{app:metrics}

For $N$ scheduled decisions with valid probabilities, we distinguish
\begin{align}
\operatorname{MSE}_q &= N^{-1}\sum_t(\hat q_t-q_t)^2,&
\operatorname{MSE}_b &= N^{-1}\sum_t(\hat b_t-b_t)^2,\\
\operatorname{Brier} &= N^{-1}\sum_t(\hat q_t-y_t)^2,& y_t&=a_t^B.
\end{align}
We use the binary, one-component convention for Brier score~\citep{brier1950verification}; constant $0.5$ gives $0.25$. Constant-baseline MSE against simulator $q_t$ varies across subsets and is not universally $0.25$. A model's sampled-action accuracy, agreement with the oracle best response, match win rate and average payoff answer different questions. With $\pm1$ rewards, a live condition's mean 50-round payoff is $50(2c-1)$, where $c$ is the fraction of favorable realized outcomes; that fraction is not forecast accuracy.

For a chosen action $a$, expected one-step regret is
\begin{equation}
\mathcal{R}_t(a)=\max_{a'\in\{0,1\}}\mathbb{E}_{Y\sim\operatorname{Bern}(q_t)}[u(a',Y)]
-\mathbb{E}_{Y\sim\operatorname{Bern}(q_t)}[u(a,Y)].
\end{equation}
The optimum depends on the focal role. Regret is calculated using the simulator probability, not a resampled outcome or a new live trajectory. Supplying true $q_t$ makes copying possible, so low error in that condition is information use rather than inference.

History-frequency baselines use Laplace smoothing on prior actions only. The $q$ baseline counts $B$'s past choices; the $b$ baseline counts $A$'s past choices. Constants and frequency predictions are rescored on the exact keys and rounds used by the comparison. Pearson correlation describes covariation, not calibration; constant predictions have undefined correlation, not zero correlation. Within-match-centered correlations remove differences in trajectory means but do not identify an internal mechanism.

\paragraph{Missing outputs and initialization.}
The R3/R5 selection and principal probability comparisons assign squared-error loss 1 to invalid or missing predictions, keeping all scheduled keys in the denominator. Conditional-valid errors and coverage are also reported. A failed own-state chain remains missing after its last valid output; no oracle reset is inserted. The GPT studies preserve both first-attempt records and explicitly linked retry-assisted views. Infrastructure failure is not imputed as an observed payoff or belief error. Initial rounds can have the trivial probability $0.5$; the main R4/R5 and GPT contrasts exclude these rounds, whereas R3's primary 600-key result includes them and reports round-wise results separately.

\paragraph{Uncertainty.}
We resample complete matches, retaining all their included rounds, and report 95\% CIs throughout. The R3 and R5 main contrasts use 5,000 paired stratified draws across the six role-by-order cells, with analysis seeds 20260916 and 20260918, respectively. New live supplement contrasts use 20,000 match-bootstrap draws (seed 20260922). Historical controls use independent draws unless an explicit paired design was frozen. Matched simulator seeds in GPT live comparisons do not make the resulting histories identical, and API sampling is unseeded. The six-match depth/memory pilot has only one trajectory per role/order cell; its MSE points are descriptive and no MSE CIs were calculated. All reported CIs are conditional on the evaluated checkpoints and prompts, not estimates of training-seed robustness. Cellwise analyses are descriptive without multiplicity correction. Several historical figure points have no comparable CI; none is invented here.

\section{Models, prompts, decoding and parsing}
\label{app:prompts}

\subsection{Models and conditions}
The scale series is Qwen2.5-Instruct at 0.5B, 1.5B, 3B, 7B, 14B, 32B and 72B. Representative other models are Qwen3.6-35B-A3B, Llama-3.3-70B-Instruct, Gemma-4-31B-it and GPT-OSS-20B; GPT-5.4 provides fresh action-only/structured live controls and separate fixed-history diagnostics. Model labels are the actual experiment identifiers, not claims about all members of each family. Figure~\ref{fig:s1} shows the baseline and flagship model/opponent/role comparisons; repeated model rows between those blocks reuse the same source cells.

The fresh open-model completion package uses 20 matches of 50 rounds per cell. Small models use one H200 (tensor parallelism 1); standard and reasoning tasks use four H200s (tensor parallelism 4). Nonreasoning baseline and structured completion budgets are 64 and 384 tokens in that package. Thinking and OSS configurations have larger frozen limits, detailed in the configuration manifests. Comparisons with older runs retain differences in runtime, tensor parallelism and sometimes prompt budgets. They are not controlled interventions on model size alone.

Fresh GPT-5.4 live comparisons use \texttt{reasoning\_effort=none} and a 2,048-token completion cap, with the same candidate-strategy glossary in both output conditions. GPT fixed-history reasoning studies use \texttt{medium}; the later complete-state, depth/memory and scalar-intervention packages request 32,768 completion tokens. Earlier 4,096/8,192-token failures are retained and not attributed automatically to intrinsic incapacity. The live GPT controls and medium-reasoning diagnoses are different experiments.

\subsection{Input and output contracts}
Natural action-only play supplies role, rewards, available binary actions and complete observable history. Structured behavioral output additionally requests the opponent strategy, its next action, the focal best action and final action. It does not require accurate numerical beliefs. The following summarizes the interfaces; exact renderers and captured journals, where available, remain authoritative.

\begin{table}[ht]
\centering\small
\caption{Information and output contracts. ``Own state'' always means previously generated state unless teacher forcing is explicitly indicated.}
\label{tab:interfaces}
\begin{tabular}{p{.21\linewidth}p{.40\linewidth}p{.28\linewidth}}
\toprule
Study & Input beyond task/role/history & Requested output \\
\midrule
{\raggedright Natural live\par} & {\raggedright None; unidentified opponent\par} & {\raggedright Action\par} \\
{\raggedright Structured live\par} & {\raggedright Candidate descriptions vary by archived prompt\par} & {\raggedright Strategy/action chain and final action\par} \\
{\raggedright Qwen token probe\par} & None & {\raggedright Digit 0/1, scored via normalized logits\par} \\
{\raggedright Qwen belief probe\par} & None & {\raggedright Decimal opponent belief $b$\par} \\
{\raggedright Probability-only SFT\par} & {\raggedright Opponent specification\par} & {\raggedright Decimal action probability $q$\par} \\
{\raggedright State-plus-probability SFT\par} & {\raggedright Specification; true prior in warmup, own prior in evaluation\par} & {\raggedright Numeric state vector and $q$\par} \\
{\raggedright Probability-only SFT continuation\par} & {\raggedright Specification on all or exactly half of training rows\par} & {\raggedright Decimal $q$\par} \\
{\raggedright GPT exact-rule\par} & {\raggedright Exact rules; history, current true state, true prior, or own state, depending on arm\par} & {\raggedright $q$; own-state arm also returns full state\par} \\
{\raggedright GPT L1/L2/L3\par} & {\raggedright No identity/rules/state; optional own-output memory\par} & {\raggedright $q$; add $b$; add a deeper nested report\par} \\
{\raggedright GPT scalar intervention\par} & {\raggedright One supplied self/true/shuffled $b$ or $q$\par} & {\raggedright Forecast and own choice\par} \\
\bottomrule
\end{tabular}
\end{table}

For example, probability-only SFT explicitly asks for JSON with a single \texttt{p\_opponent\_1} number. The state-plus-probability interface adds an ordered \texttt{state} array and identifies the last completed action pair to update from. In the matched-exposure study (R5), history-only prompts remove ToM type, parameters, coordinate definitions and public initial values together, while retaining role, payoff meanings and completed action pairs. This bundled manipulation does not isolate the effect of knowing the opponent's name.

\subsection{Reported-chain consistency does not establish causal reasoning}
\label{app:chain-consistency}
The structured live format serializes a strategy estimate, a predicted next opponent action, a focal best action and a final action. In an autoregressive response, each later field can use preceding generated text, but ordering alone is not a causal intervention on the model's internal decision. An archived, separate Qwen2.5-7B ToM2 checkpoint analysis reports that the final action matches the model's own best-action field in all 1,000 seeker and all 1,000 hider decisions per displayed condition. Yet in the untrained structured baseline, the best-action field agrees with the role-appropriate response to its own predicted opponent action in only 394/1,000 seeker decisions and 959/1,000 hider decisions. This comparison uses archived checkpoint trajectories, not the formal Figure~\ref{fig:behavior}c conditions; the mismatch cautions against equating field agreement with a faithful prediction-to-action chain. In a separate GPT-5.4 fixed-history elicitation pilot, all 54 valid outputs in each of the direct, free-summary and explicit-chain arms choose a role-consistent best response to the reported action probability. That check establishes output consistency on those observations, not live payoff or causal use of the report. The source tables and audit are \path{figures_sft_checkpoint_progress/checkpoint_progress_row_process.csv} and \path{GPT_CHAIN_RESULTS_20260917.md}; the latter records the 54-point pilot and its six-match limit.

\subsection{Parser scope and known limitations}
Strict structured parsing in the open-model package means extraction and validation of a JSON object according to the frozen parser; it does not require the entire completion to consist only of JSON. The narrowly authorized Qwen3.6 repair recovers a unique complete \texttt{final\_action} field from otherwise malformed free-text JSON. It is not a free-form semantic guess. Replaying a parser on an old response cannot retroactively replace an action and preserve the original live trajectory; corrected live conditions are therefore separate runs.

The five Qwen3.6 seeker non-ToM2 structured cells required 24 action-only repairs among 5,000 decisions, with no random fallback. Its audited ToM2 hider cell required one repair among 1,000 decisions. Gemma's ToM2 hider structured cell contains 999 strict actions and one random fallback after an incomplete response. All rounds and its resulting trajectory are retained: the score is deployed-system payoff, not performance of correctly parsed model actions alone. Exact generation finish reasons were not exported in every historical runtime, so an incomplete string is not automatically a measured token-cap exhaustion.

\section{Live benchmark audit and comparison limits}
\label{app:full-benchmark}

\begin{figure}[!t]
\centering
\includegraphics[width=\linewidth]{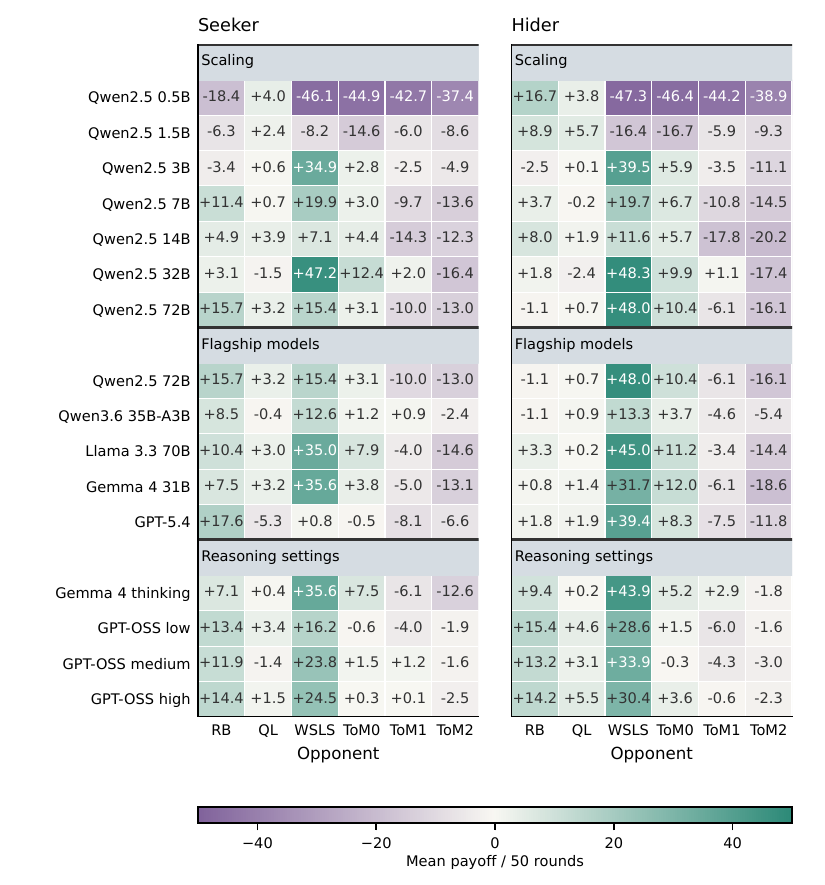}
\caption{\textbf{Live play across opponents.} Mean 50-round payoff by role, from 20 matches per cell. Purple denotes negative payoff and teal positive payoff in both panels; color does not encode role or significance. The reasoning block shows Gemma 4 thinking and GPT-OSS low, medium and high effort across all six opponents. It combines historical seeker runs with audited hider replacements, rather than paired role or effort comparisons. Prespecified random fallback actions remain in the reported payoffs. The Qwen2.5-72B row is repeated in the scaling and flagship groups using the same source cells. Qwen3.6 uses no thinking.}
\label{fig:s1}
\end{figure}

\begin{figure}[!t]
\centering
\includegraphics[width=\linewidth]{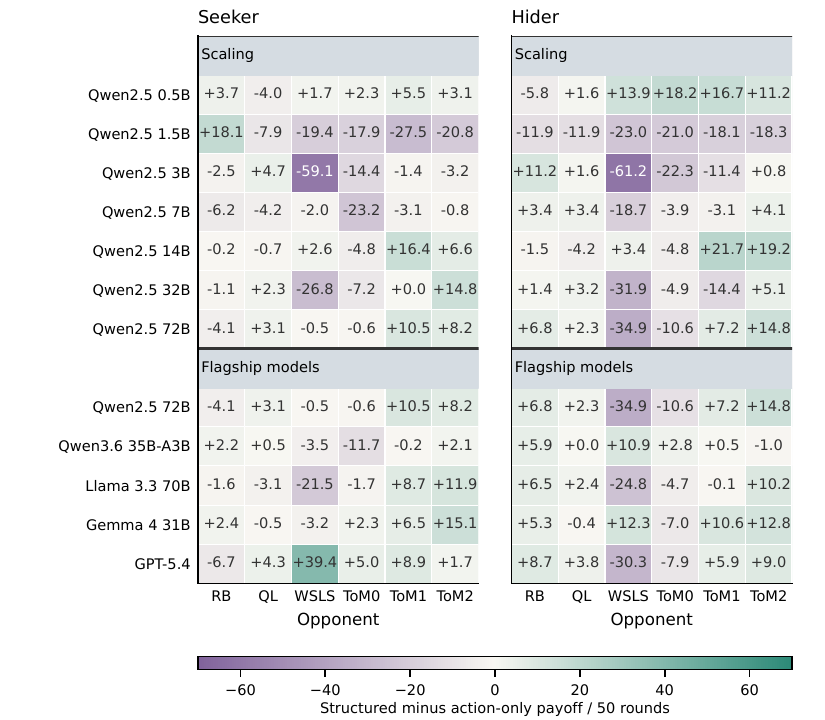}
\caption{\textbf{Structured-output payoff changes.} Each value is structured minus action-only payoff per 50-round live match; purple denotes a negative difference and teal a positive difference in both panels. Color is not a significance test. The scaling block covers seven Qwen2.5 sizes from 0.5B to 72B, both roles and all six opponents. The 72B row is repeated in the flagship group using the same contrasts. All 48 added structured-output cells have 20 complete matches; 0.5B scores retain 1,184 prespecified random fallback actions among 12,000 decisions. GPT-5.4 controls are matched by simulator seed; open-model controls are unpaired historical runs and may differ in prompt, runtime or budget.}
\label{fig:s2}
\end{figure}

Figures~\ref{fig:s1} and~\ref{fig:s2} distinguish live payoff from changes under structured output. The audited benchmark completion includes 21 small-model and 53 standard-model cells (74,000 decisions in 1,480 matches). The structured-output scaling extension adds 48 cells for Qwen2.5-0.5B, 1.5B, 7B and 14B, each with 20 matches of 50 rounds, to the existing 3B, 32B and 72B comparisons. The reasoning comparison now covers all six opponents in both roles for Gemma 4 and GPT-OSS; its hider extension adds 20 non-ToM2 cells (20,000 decisions) to the ToM2 results. The Qwen3.6 seeker ToM2 replacement and three historical Qwen2.5-7B hider baseline cells were separately audited. Fresh GPT non-ToM2 controls cover 400 matches and 20,000 decisions in addition to the ToM2 controls. These counts describe separate additions, not the total historical benchmark.

Figure~\ref{fig:s1} shows the live scores across the six opponent classes.

Figure~\ref{fig:s2} subtracts each action-only score from its structured-output score. This is a different question from whether either score is high: the sign and size of the change vary across opponents and between seeker and hider.

For each new live cell, the audit requires 1,000 unique match/round keys, 20 complete matches, role-correct rewards, past-only prompts, parsing/fallback accounting, and source/configuration/output hashes. Frozen sources must match the deployment manifest. Historical summary inclusion alone is weaker than a new semantic audit; the appendix and evidence manifest identify this distinction.

One archived 72B baseline-WSLS interval is $[48,48]$ because the 20 observed match means coincide; this does not imply zero population uncertainty. Open-model controls are unpaired historical runs with prompt, runtime and budget differences. GPT's fresh controls use matched simulator seeds but not identical realized histories. The 400-match GPT supplement includes 142 recovered matches, so it is not 20,000 uninterrupted first-attempt successes. Recovery and output-level provenance are retained in the audit records, separate from the effect estimates.

\paragraph{Reasoning results remain opponent-dependent.}
The Gemma thinking hider scores $+43.9$ against WSLS but $-1.8$ against ToM2 (95\% CI $[-4.5,0.8]$). Its six cells contain 84 prespecified random fallback actions among 6,000 decisions, including 30 against ToM2. GPT-OSS low/medium/high hider WSLS scores are $+28.6/+33.9/+30.4$; against ToM1 they are $-6.0/-4.3/-0.6$, and against ToM2 $-1.6/-3.0/-2.3$. Thus the highest effort has the highest point estimate for some opponents, but the pattern is not monotonic across the benchmark. The 15 non-ToM2 GPT-OSS hider cells retain 87 fallbacks among 15,000 decisions; the three ToM2 cells retain three/four/ten. These are independent live evaluations, not a controlled estimate of an effort effect. The Qwen3.6 seeker ToM2 structured-output replacement is audited, but its contrast uses an archived unpaired baseline.

\paragraph{Structured output has different effects across the scaling series.}
Completing the seven-size comparison does not reveal a uniform structured-output benefit. For Qwen2.5-1.5B hider, the six payoff changes range from $-23.0$ to $-11.9$ per match; for 14B hider they range from $-4.8$ to $+21.7$. Qwen2.5-7B seeker changes are negative for all six opponents, from $-23.2$ to $-0.8$. These are descriptive contrasts against historical action-only controls. The new runs use the same structured-behavior interface, 384 output tokens and 8,192 context tokens, temperature 0.7 and top-$p$ 0.9. Random fallback and action-field repair are retained in the live scores; a recovered final action does not certify that the remaining structured report is valid.

One historical hider structured prompt asks for the opponent's next ``hiding action'' even though the model is the hider; the role/payoff block is correct. Affected open-model comparisons therefore mix output format with wording differences. GPT's fresh matched controls use their own frozen prompt. Historical Gemma hider thinking results contaminated by the old parser are invalid; the six replacements instead retain native final-channel text and count their random fallbacks. Each Gemma thinking replacement first passes a ten-decision interface gate. A failed gate is not a measured payoff.

\section{Behavioral SFT and Qwen prediction studies}
\label{app:sft}
\label{app:qwen}

\subsection{Original behavioral SFT}
Both formats use histories collected from raw Qwen2.5-7B-Instruct playing RB, QL, WSLS, ToM0, ToM1, and ToM2 in both roles. Each format has 9,600 training, 1,200 validation and 1,200 test decisions, split by complete matches with role/opponent stratification. Each of the 12 cells contributes 800/100/100 decisions. The formats render the same history pool, not 19,200 independent training observations.

The action-only label is a role-conditioned best reply to the sampled current opponent action. Structured labels additionally provide the generating opponent's strategy class, sampled next action, best action and final action. They do not supervise a continuous belief $b$ or generating action probability $q$. The label builder can use simulator information for supervision without exposing it in evaluation prompts. This is behavioral SFT, not GPT distillation, online RL or direct belief-probability supervision. The figure compares structured-output step 300 and action-only step 2400 with their corresponding untrained interface baselines; those different training budgets constrain interpretation.

\begin{figure}[!t]
\centering
\includegraphics[width=\linewidth]{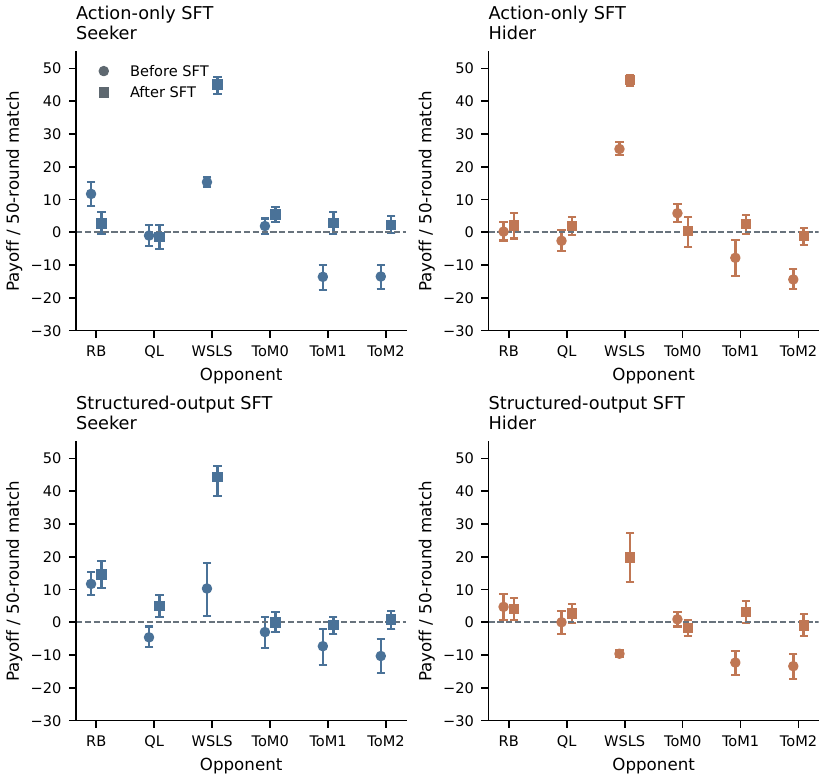}
\caption{\textbf{Behavioral SFT across six opponents and both roles.} Qwen2.5-7B, 20 live 50-round matches per cell. Blue denotes seeker and orange hider throughout; circles are the untrained interface baseline and squares the corresponding SFT checkpoint. Whiskers are archived 95\% whole-match bootstrap CIs. Action-only step 2400 and structured-output step 300 are separate training packages, each compared with its corresponding baseline; they are not equal-update conditions. These evaluations are live play, not training-history replay. ToM3 and GPT-5.4 transfer are distinct evaluations in Figure~\ref{fig:behavior}d,e.}
\label{fig:s3}
\end{figure}

Figure~\ref{fig:behavior}d,e shows the ToM2, ToM3 and GPT-5.4 live endpoints; Figure~\ref{fig:s3} shows all six training opponents. Match means and intervals remain in the plotted-value export rather than a duplicate numerical table here.

\subsection{Action and belief probes (T04)}
Four models share six frozen ten-round histories, one per role and ToM0/1/2 cell: initial Qwen, a behavioral parent, and two further-supervised checkpoints initialized from these models. Each produces 60 action probes and 60 belief reports, totaling 480 valid evaluated outputs. The behavioral parent is the structured checkpoint-300 path. Historical Figure~\ref{fig:behavior} launchers nominate that path, but the earlier runtime's loaded tensors are not independently re-attested; this is not a fully verified same-tensor payoff-to-belief causal comparison.

The additional training uses the existing 5,400/1,350 train/validation examples with five-field value supervision: a binned belief, binned action probability, opponent action, best action and final action. Ten epochs is a cap, with action-aligned validation every 100 steps, patience five and minimum improvement $10^{-4}$. Base and parent runs stop at 700 and 600 steps and select steps 200 and 100. Selection uses validation expected action Brier, not the six development histories. The action probe normalizes 0/1 token logits, whereas the belief probe requests a continuous decimal report. These are not the native five-field training interface, nor interface-matched GPT decimal forecasts.

Figure~\ref{fig:qwen}a,b,d displays the MSE contrasts: all four model action-probability and belief point estimates exceed the matched constant's errors, despite some improvement from additional supervision. The parent arm's Brier change is $-.021786$ (exploratory six-match 95\% CI $[-.045169,-.000507]$), but its selected Brier point estimate remains above constant $0.25$. All four within-match action-probability correlations are negative; this is a calibration warning, not proof of a particular failure mechanism. High 0/1 vocabulary mass (means above .99997) makes off-vocabulary mass an inadequate explanation for the token-probe errors. No MSE interval is supplied for this six-match probe. Small runtime differences relative to earlier G03 probes remain possible.

\subsection{Probability-only versus state-plus-probability SFT (R3)}
The frozen package contains 240 training, 60 validation and 60 development-test matches of ten rounds, disjoint by match ID and seed. Within each role/order cell there are 40/10/10 matches. The focal data-generating action policy is fixed before play, independently of the opponent's hidden state:
\begin{equation}
P(a_t^A=1)=\sigma\{d+1.4\sin(\phi+0.9t)\},\quad
d\sim U(-0.6,0.6),\quad \phi\sim U(0,2\pi),
\end{equation}
with generation indexed from $t=0$. This creates a controlled changing behavioral input to the opponent, not a trajectory generated by the evaluated checkpoint. Both training arms receive the specification described above and all completed pairs. Probability-only output contains $q$; state output contains the numeric vector and $q$. True previous states are used in state warmup, while common evaluation recurs on the model's own previous vector without oracle correction.

LoRA has rank 16, scale 32 and dropout .05 on attention and feed-forward projections. Both arms use AdamW, learning rate $2\times10^{-5}$, constant schedule, microbatch 1, accumulation 16, bfloat16 (BF16) and seed 2026. The per-example loss is
\begin{equation}
\mathcal{L}=\overline{\mathrm{CE}}_{q\text{ numeric}}
+0.5\overline{\mathrm{CE}}_{z\text{ numeric}}
+0.1\overline{\mathrm{CE}}_{\text{format/EOS}},
\end{equation}
where CE denotes cross-entropy and EOS the end-of-sequence token. The state term is omitted for probability-only output, and prompt loss is masked. Numerical-value token groups are normalized separately. A target-runtime gradient-accumulation regression was corrected before training; the corrected driver is hash-bound to the execution records.

Validation occurs every 50 optimizer steps with patience three, improvement threshold $10^{-4}$, and step 0 eligible. Probability-only training runs 500 steps and selects step 500. State warmup stops at 200 and selects 50; its own-state collection then generates 2,400 train-only examples. A repair stage mixes these with the original 2,400 teacher-forced examples. Its validation degrades, stops at 150, and selects repair step 0. Therefore the final state model is the warmup checkpoint, not a successful exposure-repaired model. Its larger output burden, privileged warmup states, recurrence and training time are not matched to the probability-only arm.

Figure~\ref{fig:qwen}c shows the common development-test contrast over 600 scheduled decisions per arm. The trained state-minus-probability MSE difference is $+.088240$ $[+.058681,+.128046]$. Probability-only MSE is lower than both baselines in each of the six role/order cells. The state model has two invalid outputs at round 2 in hider-ToM1 and 16 blocked descendants, completing 58/60 chains; its valid-only MSE is still $0.070243$. This deficit is therefore not explained solely by missingness. One training seed, fixed known parameters and full-history inputs preclude a claim of general state-learning impossibility or uniquely recursive computation.

\subsection{Information removal and longer histories (R4)}
R4 evaluates the selected probability-only SFT checkpoint with specification retained or removed on the same 540 noninitial keys: MSE changes from 0.01100 to 0.05024. This is an earlier cohort, not the fresh R5 cohort plotted in Figure~\ref{fig:qwen}e (0.01206 to 0.05182 at the same starting checkpoint). Both ToM2 roles in R4 have worse point estimates than constant despite partial aggregate transfer. A separate 12-match, 30-round known-information extension gives MSE .02219 versus constant .07172 and frequency .05011; on rounds 11--30, it gives .02801 versus frequency .03360. These are bounded longer-history results, not an unknown-algorithm or live-action test. Previously inspected diagnostic replay is not a new confirmatory sample.

\subsection{Matched information-exposure continuation (R5)}
Two continuations start from the same selected probability-only SFT checkpoint, with identical initial adapter tensors, 2,400 targets, row order, optimizer settings and 200 updates. The specification-only continuation retains the bundle everywhere; the mixed-exposure continuation removes it on exactly half of rows. Each ten-round training match contains five examples of each input condition. Both use the original learning rate, microbatch, accumulation and seed. Validation includes both inputs on the original 60 validation matches, scored equally over 1,080 noninitial scheduled points; step 0 and every 50 steps are candidates. Both select step 200.

Before training, a fresh 60-match development replication is frozen with ten matches per role/order cell, disjoint from training, validation and older tests. Its primary history-only result contains 540 noninitial decisions. The independent audit checks 18,720 validation/evaluation records, exact keys and prompts, 392 starting adapter tensors and 4,560 replayed simulator targets. Selected predictions are 5,520/5,520 valid across all scheduled evaluation packages; the original-checkpoint baseline has one invalid known-information response in 1,200 fresh predictions, retained with penalty 1. The primary history-only contrasts have full coverage.

The mixed-exposure continuation improves over the specification-only continuation, not merely over the old checkpoint. The specification-only versus original difference interval crosses zero. Known-information MSE is .00875 for mixed exposure and .00920 for specification-only, with a difference interval crossing zero; this is not an equivalence test. For mixed exposure versus constant on ToM2, intervals are $[-.02354,.01901]$ for seeker and $[-.01616,.00577]$ for hider. Match intervals do not measure training stability. The study changes prompt content and length together and trains $q$, not $b$; new-checkpoint live action and autonomous state maintenance remain unestablished.

\section{GPT information, depth, memory and intervention studies}
\label{app:gpt}

\subsection{Supplied-rule conditions and budget history}
All reported GPT diagnoses use full GPT-5.4, not a smaller substitute. The original matched six-history action endpoint study has history-only (H), history plus exact mechanism (K), and exact mechanism plus current true full state (S) conditions. At an 8,192-token budget, coverage is 60/60, 23/60 and 59/60, respectively. True-state conditional-valid action-probability MSE is $4.86\times10^{-10}$. The 37 length failures and one API timeout remain distinct from measured prediction errors.

A true-previous-state single-update study (U) tests 54 noninitial inputs, with rules and the most recent observation. At 8,192 tokens, 24 responses are valid and near-exact; 30 are length-terminated. A selected-failure 32,768-token follow-up returns 24 valid responses out of 30 attempts, 23 within the stated near-exact criterion, with six connection errors and no length terminations. It is not a uniform larger-budget rerun. These controls rule out treating the earlier budget failures as evidence of intrinsic update inability.

At 32,768 requested tokens, K returns 60/60 valid responses and 49/60 action probabilities within .001; excluding initialization gives 43/54. The own-state recursive condition (R) supplies exact rules, public initialization, the latest observed pair and the model's previous full-state output. It returns 58/60 scheduled valid predictions, all within .001; noninitial coverage is 52/54. Five of six ten-round chains finish. Seeker-ToM2 terminates at round 9 and round 10 is protocol-blocked. The reported finish reason is length, but usage is 8,285 against a requested 32,768 cap, so actual cap exhaustion is not established. State-coordinate accuracy is weaker than action readout: 46/54 noninitial scheduled states meet the uniform .001 criterion. Accurate $q$ therefore does not certify every internal coordinate. K and R differ in decomposition and output requirements, not just memory.

\subsection{Autonomous elicitation and same-observation controls}
The initial explicit-chain pilot compares direct forecasting, a free summary and a belief chain on 54 noninitial points from six matches. Forecast MSE is .04990, .05747 and .04097, respectively. Chain belief MSE .02723 is worse than constant .01169 and frequency .01353. These descriptive six-match results motivate a larger, separately generated development replication, not a claim of robust chain superiority.

The replication uses 24 additional ten-round matches, four per role/order cell, with 216 noninitial points per arm. L1 reports $q$; L2 additionally reports $b$. Both see history but no identity, parameters, update rules, current state or carried prior reports. A separate supervised same-observation baseline is trained on independent trajectories, supplying an attainability control with an additional task-specific learned prior. It is not an equal-training comparison with zero-shot GPT.

The separately trained same-observation control has forecast MSE $0.01917$ versus GPT L2's $0.05910$, but it has additional task-specific experience. L2-minus-L1 regret is $-.00667$ with 95\% match interval $[-.07386,.06031]$. Belief-report correlation can coexist with worse absolute error than constant; it does not resolve the role/order differences shown in Figure~\ref{fig:gpt}b,c.

\subsection{Report depth and self-generated memory}
The depth experiment crosses three output requirements with two memory conditions on six ten-round histories: L1 reports $q$, L2 adds $b$, and L3 adds a deeper nested belief. No-memory calls receive the observable history alone; memory calls additionally retain the model's actual earlier outputs. They never receive oracle corrections. The deepest scalar lacks a validated mapping to a unique simulator target and is not scored as recovered deep belief. The factorial design has 360 scheduled outputs across 36 chains.

Figure~\ref{fig:gpt}f displays each depth-by-memory forecast MSE relative to the matched constant, with no duplicated matrix here. L2 with memory has lower aggregate forecast MSE than L1 with memory (.03491 versus .04157), but its ToM2 average combines different seeker and hider outcomes. With only one trajectory per role/order cell, these descriptive MSE differences do not establish a reliable role-general benefit; no MSE CI was calculated. L3-with-memory belief MSE is .02767 versus constant .01169, so a lower forecast error does not establish accurate belief recovery.

The original first-attempt view has 339 valid outputs, seven length failures and 14 blocked descendants. Authorized same-configuration retries recover seven failures and continue the 14 descendants from actual valid own outputs, giving 360/360 in the retry-assisted view. Original outputs and failure records remain unchanged. The extra-attempt policy must not be represented as 360 first-attempt successes.

\subsection{Matched external-information intervention}
On 54 noninitial decisions from six matches, six input arms supply self-generated, true or shuffled values of either $b$ or $q$. Prompts request a forecast and focal action. True values come from decision-time simulator state; shuffled values are drawn from other fixed histories according to the frozen donor mapping. Donors need not be reachable under the recipient history, and changing a supplied value is not an intervention on the model's internal neural state.

In the same audited decisions, the final action agrees with the role-conditioned reference best action in 36/54 self-$b$ calls and 52/54 true-$b$ calls. This agreement is tabulated from the action records; Figure~\ref{fig:gpt}d plots forecast MSE and (e) plots expected-regret differences, neither of which directly plots agreement.

Beyond the main-text true-$b$ versus self-$b$ comparison, true-minus-shuffled $b$ regret is $-.29640$ $[-.39106,-.19726]$, and true-$q$ versus self-$q$ regret is $-.10584$ $[-.15193,-.06614]$. Each arm uses 54 fixed-history decisions from six matches, not 54 independent matches or a live payoff test. These effects of correct external information do not require complete recursive inference for a good choice. All six arm-level values remain in the audited source export.

Together, intervention and replication schedule 804 outputs. Initially 791 are valid and 13 return HTTP 500 errors. Thirteen explicitly authorized unchanged-configuration retries give 804/804 unique valid keys after 817 issued attempts. All 791 original valid outputs and original failures are retained; conditional-valid and penalized first-attempt analyses remain separate. Retry-assisted completion is not a new independent replication.

\section{Historical diagnosis, RL and live predictive deployment}
\label{app:history}
\label{app:audits}
\label{app:closed-loop}

\revold{These earlier studies use heterogeneous interfaces and are context, not additional arms of the main factorial comparisons. Archived Qwen32B and 72B ToM2 offline opponent-action accuracies (.597 and .639) exceed their online action accuracies (.334 and .370), but the endpoints differ and this is not a matched causal gap. Historical GRPO~\citep{shao2024deepseekmath} and online-refresh variants improve some simple-opponent play, with non-monotonic checkpoints and role-dependent transfer; neither certifies a transferable recursive state.}

\revold{The archived 600-decision recursive-GRPO prefix test finds reference-best-action agreement .6350 with free output, .6383 with either true or shuffled belief, and .8250 with true action probability. True and shuffled belief nevertheless change 62/600 individual actions, so equal means do not imply no field use; the forecast-MSE specificity interval crosses zero. This contrasts with, but does not override, the separate GPT supplied-information result. In another historical path, realized-action SFT improves OOD Brier from .347952 to .228034 on 2,626 observations, while a same-feature ExtraTrees control is better at approximately .2194. A 60-match-per-model live ToM3--5 test improves mean payoff from $-9.4333$ to $+.2667$ using an external probability-to-action map. It is neither new R3/R5 live closure nor native execution of the full belief-to-action chain.}

\begingroup
\ifrevisionmarkup\color{blue!70!black}\fi
\subsection{RL training families and evidence scope}
\label{app:rl}

These exploratory GRPO~\citep{shao2024deepseekmath} studies test fixed-buffer action-chain training, policy-refreshed training, and continuous-probability rewards. Different data, objectives, group sizes and evaluation cohorts preclude an equal-compute algorithm comparison or pooling with the main experiments.

Preserved configurations initialize Qwen2.5-7B-Instruct from structured-output behavioral SFT step 300: LoRA rank 16, scaling 32, dropout .05, learning rate $3\times10^{-6}$, Kullback--Leibler (KL) coefficient $10^{-4}$, temperature .9 and seed 2026, adapting attention and multilayer-perceptron (MLP) projections. Offline training uses six completions per prompt ($K=6$), 600 updates, and prompt/completion limits of 8,192/768 tokens. Training has 9,600 examples from 192 matches across ToM0/1/2, RB, QL and WSLS in both roles; validation has 1,200 examples from 24 disjoint matches. Reward assigns .05 to valid format, .95 to correct realized opponent-action prediction, .5 to correct role-conditioned best action conditional on correct prediction, and .5 to correct final action conditional on preceding fields. This scores an action chain, not continuous beliefs or match return. Labels enter reward, not evaluated prompts. Training requests brief reasoning and action fields; evaluation uses the structured-output interface.

Evaluation launchers default to temperature .7, top-$p$ .9, context 8,192 and local completion limit 384; historical overrides were not recovered. We revalidate archived keys, coverage and payoffs where raw exports exist, without re-attesting original training tensors or model revisions. Checkpoints were repeatedly inspected without an independent confirmation set. Intervals quantify match variation, not training-seed or selection uncertainty. Full checkpoint tables remain in the evidence records.

\subsection{Fixed-buffer action-chain GRPO}

Figure~\ref{fig:rl-offline} gives all nine algorithmic opponents plus GPT-5.4 at the three archived checkpoints. Each role/opponent/checkpoint cell contains 20 live matches of 50 rounds. This archive has no matched step-zero evaluation, so the figure shows checkpoint performance and trends within RL, not a causal pre/post-RL effect. All 60,000 algorithmic/GPT evaluation rows have unique cell/match/round keys and payoffs consistent with executed actions.

\begin{figure}[!t]
\ifrevisionmarkup\color{blue!70!black}\fi
\centering
\includegraphics[width=\linewidth]{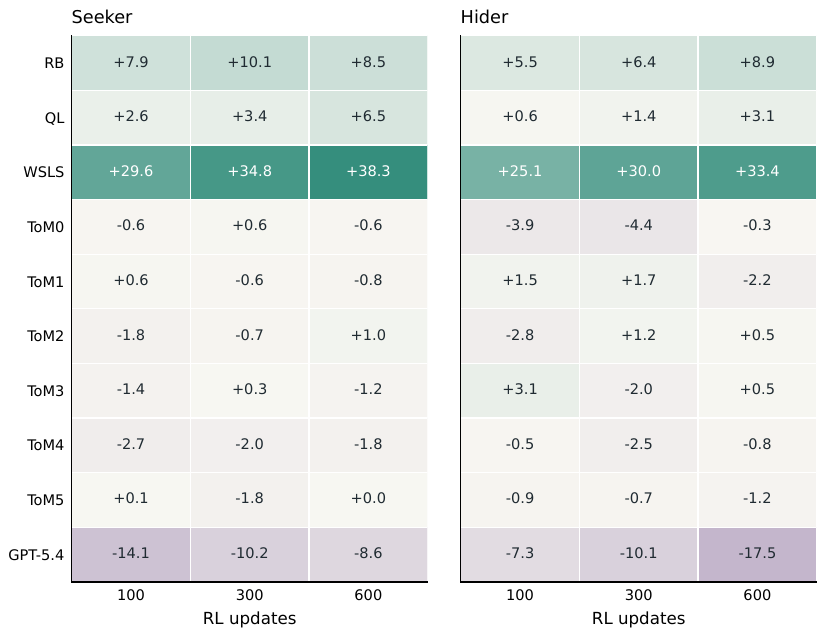}
\caption{\textbf{Offline action-chain GRPO across opponents.} Each cell gives mean payoff per 50-round live match, with 20 matches per role and checkpoint. Teal denotes positive payoff and purple negative payoff in both role panels; numbers give the exact archived means. All nine algorithmic opponents and GPT-5.4 are retained. No matched pre-RL checkpoint or uniform uncertainty layer is available, so color differences are descriptive, not a significance test.}
\label{fig:rl-offline}
\end{figure}

The clearest checkpoint trend is stronger WSLS play. ToM outcomes are smaller and non-monotonic; GPT transfer improves descriptively for the seeker but worsens for the hider. At step 600, archived 95\% match-bootstrap payoff intervals against GPT-5.4 are $[-12.4,-4.7]$ and $[-20.3,-15.0]$ for seeker and hider. Separate self-play evaluations (20 two-role matches per checkpoint) give seeker payoffs $11.1$, $18.3$ and $2.9$; hider payoffs are their negatives. This imbalance describes role asymmetry, not independent success for both players or recovery of a recursive state.

\subsection{Policy-refreshed action-chain GRPO}

A distinct one-cycle self-play run collects 20 50-round matches per candidate role against a frozen SFT copy, then performs 300 updates with $K=4$ using the action-chain reward. It does not refresh trajectories after each update. Its final checkpoint earns $26.9/27.2$ against WSLS, $-1.5/-.3$ against ToM2, and $-11.9/-10.4$ against GPT-5.4 (seeker/hider, 20 matches per cell); self-play is $+5.5/-5.5$. Its six-checkpoint sweep is non-monotonic. This is a different trained model from the offline $K=6$ run, not a repeat of that evaluation.

Later iterative training collects current-policy-versus-frozen-policy trajectories before each update block, using the preceding adapter as the next frozen opponent. Three iterations of 75 updates give 225 cumulative updates; each refresh collects four 50-round matches per role. The action-chain reward is multiplied by a capped role-asymmetry weight. The pilot uses $K=4$, confirmed by saved completion groups; an older narrative incorrectly called it $K=6$. The larger-group variant uses $K=16$.

Both variants evaluate ToM2 with 20 matches per role/checkpoint. Figure~\ref{fig:rl-online} reports the SFT start and each iteration endpoint. Intermediate points also fluctuate: mean-role payoff reaches $.35$ at step 100 then $-.95$ at step 125 for $K=4$, and $-1.55$ at step 125 for $K=16$. At the $K=16$ terminal checkpoint, seeker payoff is $-.5$ (95\% CI $[-3.8,2.9]$) and hider payoff $+2.0$ ($[-.2,4.2]$). The mean $.75$ is also the best inspected checkpoint, not an independently confirmed selection. Self-play ends at $+17/-17$ seeker/hider, versus $+6.8/-6.8$ at the SFT start. Neither series establishes a reliable benefit of additional updates or a larger group.

\begin{figure}[!t]
\ifrevisionmarkup\color{blue!70!black}\fi
\centering
\includegraphics[width=\linewidth]{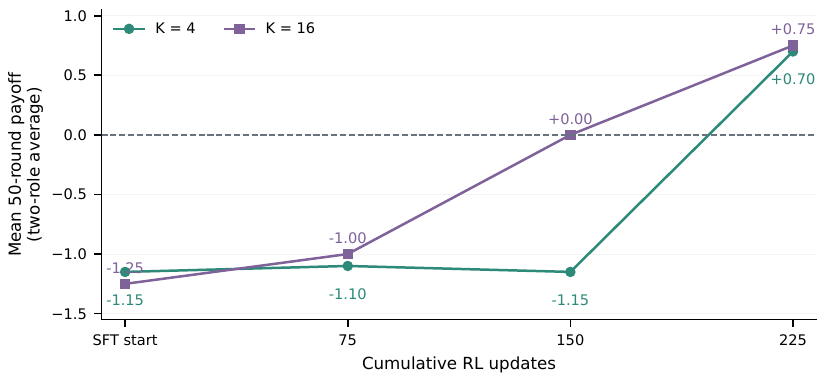}
\caption{\textbf{Iterative GRPO against ToM2.} Points are mean 50-round payoff averaged across seeker and hider, each evaluated in 20 matches at the SFT start and each iteration endpoint. Teal circles and purple squares distinguish $K=4$ and $K=16$; color does not encode role. Lines join checkpoints within a training run; evaluation cohorts differ between group sizes. No comparable confidence interval is available for every point, and these curves do not establish a reliable update or group-size effect.}
\label{fig:rl-online}
\end{figure}

\subsection{Probability-targeted and recursive RL}
\label{app:rl-probability}

These variants explicitly reward simulator targets and must not be described as payoff-only RL. Single-probability GRPO rewards $.25-(\hat q-q)^2$, .05 for valid format, and .05 each for probability-to-declared-action, declared-action-to-role-best-response, and best-to-final consistency; invalid output receives $-.75$. It does not supervise $b$. The early checkpoint series includes adverse cells: seeker-ToM0 at 200 updates earns $-26.3$ payoff, with target-$q$ MSE $.4302$ and realized-action Brier $.5122$, despite best-to-final consistency of 1.0.

Online-buffer variants use five 10-update or three 30-update cycles. A separate longer gate starts again from the SFT parent and performs one 120-update cycle with $K=16$ on newly collected data. Its 600 decisions are split by whole match into 300 training and 300 validation rows; the buffer is not refreshed during those 120 updates. Across 120 final live matches (20 per role and ToM0/1/2 cell), target-$q$ MSE is $.11749$, worse than constant $.5$ at $.07783$, and correlation with the target is $-.209$. Mean match payoff is $-4.37$. Yet the three consecutive output-consistency rates are $.9945$, $.9905$ and $1.0000$. Correctly relating reported fields therefore coexists with inaccurate input to that chain. Training/evaluation target distributions differ, but these observations alone do not identify the cause.

The recursive-probability v4 smoke adds a weak belief squared-error term, action-probability calibration, confidence-weighted direction rewards and gated expected-action regret. After 25 updates with $K=16$, archived matched summaries report seeker payoff changing from $-13.67$ to $-4.33$ while opponent-action prediction accuracy falls from $.6967$ to $.6033$; hider payoff changes from $+.67$ to $-6.00$. Each role has only six matches, two per ToM0/1/2 opponent. Post-training belief MSE is $.10575$ versus constant $.02602$ for seeker and $.06184$ versus $.02206$ for hider. These are small-sample summary-level findings: the complete original v4 generation/configuration snapshot was not re-attested. They show why better payoff in one role should not be relabeled as belief recovery.

\clearpage
\subsection{Does a trained model selectively use a correct belief?}
\label{app:rl-prefix}

\begin{figure}[!h]
\ifrevisionmarkup\color{blue!70!black}\fi
\centering
\includegraphics[width=\linewidth]{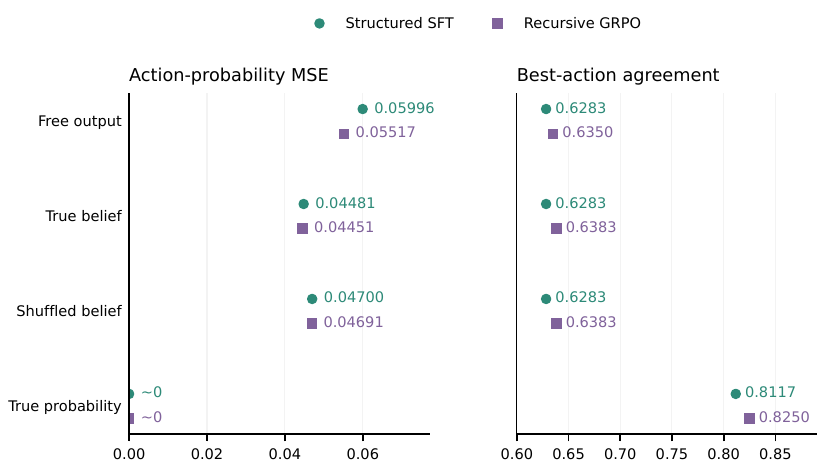}
\caption{\textbf{External-prefix diagnostic on fixed histories.} Each arm has 600 decisions from 12 histories. Teal circles identify structured SFT and purple squares recursive GRPO; color does not encode role. Left: action-probability MSE (lower is better); right: agreement with the reference best action (higher is better), not live win rate. Text labels retain the archived point estimates. Supplying true $q$ fixes that scored field, so its near-zero MSE is not a learned forecast. No arm-level uncertainty is plotted; the paired true-versus-shuffled-belief interval is reported in the text.}
\label{fig:rl-prefix}
\end{figure}

A separate archived prefix experiment compares a structured-output behavioral SFT checkpoint and a 25-update recursive-GRPO checkpoint on 12 fixed histories, 600 decisions per arm. The recorded adapter path identifies the earlier recursive-v3 checkpoint, distinct from the v4 smoke above. Figure~\ref{fig:rl-prefix} maps the archive's opponent-belief field to $b$ and opponent-action-probability field to $q$. All arms share match/round keys and simulator targets. Agreement is with the reference best action, not live win rate.

For recursive GRPO, shuffled-minus-true-$b$ forecast MSE is $.00240$, with a newly recomputed paired whole-match bootstrap 95\% CI $[-.00119,.00625]$ (20,000 resamples). No clear advantage is detected, but this is not equivalence. True and shuffled $b$ change 62/600 final actions (64/600 for SFT), despite equal mean action agreement. Donors come from different matches; 32/600 beliefs are numerically almost identical to the true value. Donor states need not be reachable in the recipient history. Prefixing also skips generation of that field, and the archive lacks a format-matched self-belief prefix and irrelevant-field control. It therefore does not isolate all gains over free output to belief correctness. This contrasts with the separate GPT supplied-belief study without establishing a matched model or training effect.
\endgroup

\revnew{\subsection{Other historical prediction and deployment diagnostics}
Archived Qwen32B and 72B ToM2 offline opponent-action accuracies (.597 and .639) exceed their online action accuracies (.334 and .370), but the endpoints differ and this is not a matched causal gap. In a separate realized-opponent-action prediction SFT path, out-of-distribution (OOD) Brier improves from .347952 to .228034 on 2,626 observations; a same-feature ExtraTrees control is better at approximately .2194. This path predicts a sampled opponent action, not the focal action labels of behavioral SFT. A 60-match-per-model live ToM3/4/5 test improves mean payoff from $-9.4333$ to $+.2667$ using an external probability-to-action map. This is partial predictive deployment evidence, not live validation of the newer probability checkpoints or native execution of the full belief-to-action chain.}

A full-history checkpoint once described as trained on model-generated priors actually used oracle priors after an export overwrote a prediction field. It remains labeled oracle-prior-trained. The corrected own-prior run has live Brier .27141 and payoff per round $-.00222$ over 180 ten-round matches; supplying the true prior lowers fixed-observation Brier to .23470 versus .27141 for own-prior recursion. This is partial privileged rescue, not autonomous belief tracking. The historical ten-attempt ledger includes changing supervision and coverage, so a failed composite gate is not ten independent tests of one mechanism. Full records, including negative outcomes and the provenance correction, remain in the experiment index.

\section{Artifact provenance and reproducibility}
\label{app:artifacts}

The code and data repository includes numerical source tables and a manifest of source and released-file hashes, including figure hashes. Audited squared-error fields use MSE for forecast and belief panels.

\paragraph{Execution and audit are different evidence levels.}
Recent Qwen/GPT records verify unique keys, original output provenance, information boundaries, target replay, parsing and selected checkpoints to the stated scope. Some original model revision or weight hashes were not recorded, and historical endpoint tensors were not independently re-attested. The R3 prompts are reconstructed from hash-bound source and prior outputs, not universally captured verbatim at runtime. R5 target replay uses the disclosed $10^{-8}$ tolerance across environments (maximum observed difference $2.18\times10^{-9}$). Such limitations are not removed by packaging the figures.

\paragraph{Compute and release.}
Recent R3/R5 training uses one H200 per branch; newer standard/reasoning live evaluation uses four H200s, while small-model evaluation uses one. Requested resource caps are not measured consumption. The two R5 jobs allocate 29m17s and 27m09s; these durations describe those jobs, not total project-wide compute. The public code and data repository linked above provides selected outputs, training and validation examples, and evaluated adapters. Its documentation distinguishes numerical reanalysis, checkpoint reconstruction and model-based reruns.

Single-seed training, small diagnostic cohorts and heterogeneous interfaces bound the claims above.

\end{document}